\documentclass[11pt]{article}

\usepackage[margin=1in]{geometry}
\usepackage{booktabs}
\usepackage{siunitx}
\usepackage{graphicx}
\usepackage{tikz}
\usetikzlibrary{positioning, arrows.meta, shapes.geometric, fit, backgrounds, calc}
\usepackage{hyperref}
\usepackage{cleveref}
\usepackage{microtype}
\usepackage{xcolor}
\usepackage{listings}
\usepackage{orcidlink}
\title{Streamlined Constraint Reasoning via CNN Pattern
Recognition on Enumerated Solutions}
\author{Patrick Spracklen~\orcidlink{0009-0007-4409-5338}}
\date{\today}

\begin{document}
\maketitle

\begin{abstract}
Constraint programming practitioners accelerate hard problems
through a layered set of techniques applied in order of risk.
Standard hardening (symmetry-breaking constraints and implied
constraints) preserves satisfiability and is applied first.
\emph{Streamliner constraints}, which restrict search to a
structural sub-family of solutions, do not preserve satisfiability
and are reserved as a final lever for instances where hardening
alone is insufficient.  Existing automated streamliner-synthesis
approaches either search a structured constraint grammar with
combinatorial budgets or prompt a Large Language Model directly on
the problem model.  We propose a different approach: enumerate
feasible solutions to a training instance, train a Convolutional
Neural Network contrastively against perturbed non-solutions to
detect structural patterns the feasibility predicate exploits,
and translate the CNN's discriminative signal, via both
filter-property correlations and direct filter-contrast
between high-activation and low-activation solutions, into candidate
MiniZinc streamliners through LLM-driven constraint synthesis.
The CNN serves as a pattern recognizer over enumerated solutions,
surfacing structural regularities that ground the LLM's
constraint-generation reasoning.
We evaluate on hardened variants of standard benchmark models
(symmetry-breaking and implied constraints already in place),
where streamliner discovery is the residual performance lever.
Our pipeline achieves \num{98.8}\% portfolio time reduction on
hardened Vessel Loading, \num{98.6}\% on hardened Social
Golfers, and \num{89.4}\% on Black Hole, with best-single
streamliners reaching geometric-mean speedups of
\num{932}$\times$, \num{356}$\times$, and \num{1103}$\times$
respectively.  Discovered streamliners include class-based
packing constraints on Vessel Loading, beyond-hardening
canonicalisations on Social Golfers, and layout-coordinate
bounds on Black Hole: structural restrictions that the
standard hardening does not provide.
\end{abstract}

\section{Introduction}\label{sec:intro}

Constraint programming (CP)~\cite{rossi2006handbook} solves
decision and optimisation problems by propagating declarative
constraints over finite-domain variables. Practitioners
accelerate hard problem instances through a layered set of
techniques applied in order of risk to the underlying problem
specification.  \emph{Symmetry-breaking constraints}~\cite{crawford1996symbreaking},
which fix a canonical representative in each orbit of the
problem's symmetry group, and \emph{implied
constraints}~\cite{charnley2006implied}, which state
consequences of the formulation that aid propagation, both
preserve satisfiability of the original instance:
implied constraints leave the solution set unchanged, while
symmetry-breaking constraints retain at least one
representative from every orbit so a model is
satisfiable iff the original is.  These sound techniques are
applied first, and form the standard hardening of any
well-engineered CP model.

\emph{Streamliner constraints}~\cite{gomes2004streamlined} sit
outside this set.  A streamliner restricts search to solutions
sharing a particular structural property, pruning large portions
of the search tree at the cost of completeness: a streamlined
instance may be rendered unsatisfiable even when the original has
solutions, and proving unsatisfiability on a tightly-constrained
streamlined instance can itself be expensive.  Because the satisfiability tradeoff is real, a
streamliner is properly viewed as a tool of \emph{last resort}:
the CP practitioner should harden the model with sound
techniques first, and only invoke a streamliner when those
options have been exhausted on instances that remain
intractable.  This ordering matters for evaluating automated
streamliner-synthesis methods: a method evaluated on an
unhardened model may be measuring the recovery of constraints a
practitioner would have added by hand before contemplating a
streamliner.

\paragraph{Prior automated approaches.}
Two families of automated streamliner-synthesis methods have
been published.  The first conducts directed search over a
structured constraint
grammar~\cite{wetter2015streamlined,spracklen2018montecarlo,spracklen2019optimisation,spracklen2023portfolios}:
candidate streamliners are drawn from a constraint lattice with
Monte-Carlo or portfolio-based selection.  The approach is
sound and reproducible but the candidate space is bounded by
the grammar the authors anticipated; constraint shapes outside
the grammar cannot be discovered.  The most recent approach,
StreamLLM~\cite{voboril2024streamllm}, prompts a Large Language
Model with the MiniZinc model and asks it to emit candidate
streamliners with closed-loop runtime feedback.  This removes
the grammar bound and reports large speedups across a
benchmark suite.  The cost is that the LLM is working from
parametric knowledge alone: it knows what constraints look like
for this problem class, but has no grounding in what structural
patterns actually hold in the feasible solutions of the specific
instance at hand.

\paragraph{The hardening question.}
Both prior approaches are evaluated against benchmark models in
their as-shipped form, which for several CP problems omits
literature-standard hardening.  Our own measurements
(\Cref{sec:hardening}) make the consequences concrete: adding
literature-standard symmetry-breaking and implied constraints to
the StreamLLM-distributed Hypergraph Coloring and Vessel
Loading models reduces the median Chuffed baseline solve time
from $\geq$\SI{1800}{\second} to under \SI{20}{\second}, with
Hypergraph Coloring collapsing to \SI{0.2}{\second}.  Streamliner-discovery results obtained
against these unhardened baselines therefore conflate two
different signals: the recovery of missing model constraints,
and the discovery of genuine streamliners.  The harder and more
practical question is what remains \emph{after} the standard
hardening is applied, and whether useful streamliners can
still be found in that regime.

\paragraph{Our approach: pattern recognition on enumerated solutions.}
Rather than searching a constraint space, we extract structure
directly from observed solutions and translate that structure
into candidate streamliners.  For each training instance we
enumerate at least \num{500} feasible solutions with a complete
solver and train a 3-layer Convolutional Neural Network
contrastively against perturbed non-solutions.  The CNN's
filters learn to detect structural patterns the feasibility
predicate exploits.  We surface the CNN's signal in two
complementary ways: (i) by correlating each filter's
activations with a problem-shape-specific library of
$\approx 30$ structural properties (row and column aggregates,
adjacent-pair statistics, positional centroids, and so on),
producing a relevance ranking we present to the LLM alongside
per-property summary statistics; and (ii) by directly
comparing high-activation against low-activation solutions
under top-variance filters, asking the LLM to hypothesise the
structural shape each filter is detecting.  Both paths emit
candidate MiniZinc streamliners that are pooled, semantically
clustered, and validated against cached baselines.

The structuring choice is that the CNN serves as a
\emph{pattern recognizer over enumerated solutions}: it
extracts structure from the data without needing a
hand-written constraint grammar, and that structure grounds
the LLM's constraint synthesis in observable evidence rather
than text-level inference from the problem model.

\paragraph{Evaluation regime and headline results.}
We evaluate on hardened benchmark models, where the standard
sound techniques have been applied and only genuine streamliner
discovery remains as a performance lever.  All headline
portfolio numbers below use the family-budget allocation
(\Cref{sec:pipeline-portfolio}) at $k = 3$ families with
$m = 3$ members per family; we additionally report simple
top-3 numbers as a comparison baseline.  Our pipeline delivers \num{98.8}\% portfolio time reduction on
hardened Vessel Loading, \num{98.6}\% on hardened Social
Golfers, and \num{89.4}\% on Black Hole (the one benchmark
in our suite without canonical hardening, making the
head-to-head comparison with prior work methodologically
clean), matching StreamLLM's realtime result on Black Hole with
simple-top-3 and exceeding it with the family-budget
allocation; the discovered constraints are qualitatively
distinct from those found by model-text-only synthesis
(\Cref{sec:eval-bh}).  Best-single
streamliners reach geometric-mean speedups in the hundreds-to-
thousands range on the instances they retain; per-problem
numbers and retention denominators are reported in
\Cref{sec:eval}.  The discovered constraints are structurally
distinct from what the hardening provides: class-based
packing constraints on Vessel Loading, beyond-hardening
canonicalisations on Social Golfers, and layout-coordinate
bounds on Black Hole.  This gives direct evidence that the
pipeline finds residual structure that the standard sound
techniques do not capture.

\paragraph{Contributions.}
\begin{enumerate}
\item A streamliner-synthesis pipeline driven by CNN pattern
recognition on enumerated solutions, with two LLM-prompting
paths (filter-property correlation and direct filter-contrast
on solution pairs) that together turn observed solution
structure into candidate MiniZinc constraints
(\Cref{sec:pipeline}).
\item Empirical demonstration that this pipeline finds useful
streamliners both on hardened benchmarks
(\Cref{sec:eval-hardened-sg,sec:eval-hardened-vl}) and on
Black Hole, the one benchmark in our suite without canonical
hardening (\Cref{sec:eval-bh}).  On Black Hole, where direct
comparison with StreamLLM~\cite{voboril2024streamllm} is
methodologically clean, the pipeline matches their realtime
result with simple-top-3 and improves on it with family-budget
allocation; crucially, the discovered constraints are
qualitatively distinct from those the model-text-only approach
finds, illustrating the complementary nature of the two
pipelines (\Cref{sec:eval-bh}).
\item Empirical evidence that adding literature-standard
hardening to three StreamLLM benchmark models
collapses median solver time from
$\geq$\SI{1800}{\second} to under \SI{20}{\second}
(\Cref{sec:hardening}), motivating evaluation of
streamliner-synthesis methods only on hardened baselines
where streamliner discovery is the residual lever rather
than recovery of missing model constraints.
\item A \emph{family-budget portfolio allocation}
mechanism that uses LLM-generated constraint descriptors as
semantic family tags to enforce both inter-family and
intra-family diversity in the deployed portfolio
(\Cref{sec:pipeline-portfolio}).  At
$k = 3$ families with $m = 3$ members per family, the
mechanism improves over simple top-3 selection on every
problem in our suite without any increase in compute envelope;
the largest gain is on Black Hole, where the simple-top-3
portfolio is dominated by redundant pile-top variants
(\Cref{sec:analysis-selection}).
\end{enumerate}

\section{Related Work}\label{sec:related}

\paragraph{Streamlining and its automation.}
Gomes and Sellmann introduced streamlining as a principled
incomplete-search technique for constraint satisfaction
problems~\cite{gomes2004streamlined}: the search is restricted
to a structurally typical sub-family of solutions in exchange for
substantially reduced search effort.  The technique has been
extended to constructive procedures
\cite{lebras2012streamlined} and applied with manual designs to
spatially-balanced Latin squares
\cite{smith2005streamlining}, with striking results on individual
problem families but expensive replication across new problems.
This motivated work on automated streamliner synthesis.
Wetter, Akg{\"u}n, and Miguel~\cite{wetter2015streamlined}
introduced a constraint-lattice search over the Essence
modelling language~\cite{frisch2008essence} via the Conjure
constraint modeller~\cite{akgun2022conjure}, and Spracklen
\emph{et al.}~\cite{spracklen2018montecarlo,spracklen2019optimisation,spracklen2023portfolios}
extended this with Monte-Carlo Tree Search, optimisation-problem
support, and portfolio-based instance selection.
The approach is sound and the portfolio framework gives strong
empirical results, but the candidate space is fundamentally
bounded by the grammar: constraint shapes that fall outside the
authors' atomic vocabulary cannot be discovered.  Our work
extends the candidate space by extracting structural patterns
directly from observed solutions and translating them into
candidate constraints via a Large Language Model, removing the
grammar bound while retaining the validation-driven filtering
of the portfolio approach.

The most recent work in this area is
StreamLLM~\cite{voboril2024streamllm}: a Large Language Model
is prompted with the MiniZinc problem model and asked to emit
candidate streamliners with closed-loop runtime feedback,
removing the grammar bound that the bottom-up methods carry.
The authors are methodologically transparent about constraint
types: their Table~1 classifies each winning constraint as
implied (\texttt{i}), symmetry-breaking (\texttt{s}),
satisfaction-non-preserving (\texttt{-}), or uncertain
(\texttt{u}); and they observe that the one benchmark model
retaining its standard symmetry breaking (BIBD) yields more
modest improvements than the others.  Our work differs from
theirs in two respects.  First, the LLM in our pipeline
reasons about \emph{observed solution structure} via a CNN
pattern recognizer rather than the model text alone.  Second,
we evaluate the synthesis pipeline on benchmark models with
literature-standard hardening already applied, isolating the
streamliner-discovery contribution from the recovery of
missing model constraints.

\paragraph{Constraint acquisition.}
A separate research area learns explicit constraint models
from example solutions.  ModelSeeker~\cite{beldiceanu2012modelseeker}
performs algebraic matching against a global-constraint
catalogue to extract per-relation constraints from positive
examples; subsequent constraint-acquisition systems use
similar template-driven matching against fixed catalogues~\cite{bessiere2017acquisition}.
These systems target the construction of \emph{satisfiability-preserving}
models from data and are template-bounded by their target
constraint catalogue.  Our pipeline targets a different
artefact (streamliners, which are explicitly
satisfiability-non-preserving) and does not require a target
catalogue: candidate constraints are generated by the LLM
conditioned on observed solution structure, not matched against
a known set of patterns.

\paragraph{Deep learning and language models for combinatorial reasoning.}
A separate line of work applies deep learning directly to
combinatorial reasoning.  SATNet~\cite{wang2019satnet} and
NeuroSAT~\cite{selsam2019neurosat} learn end-to-end neural
solvers for SAT.  Neither extracts human-readable constraints
or plugs back into a CP solver, so the systems are
complementary to ours rather than competing.  More recent
LLM-driven approaches share our pattern of using a language
model to propose candidate formal artefacts validated by a
sound downstream procedure: FunSearch~\cite{romera2023funsearch}
generates candidate program fragments evaluated against an
automated scorer for combinatorial discovery; Lemur
\cite{wu2023lemur} integrates LLMs into automated program
verification; and Pei \emph{et al.}~\cite{pei2023invariants}
ask whether LLMs can hypothesise loop invariants, with the
hypotheses checked by a verifier.  Our pipeline shares this
generate-and-validate structure but applies it to streamliner
synthesis for CP, with the additional ingredient of CNN-based
pattern recognition over enumerated solutions providing the
LLM's evidential grounding.

\paragraph{Feature interpretation in CNNs.}
Our use of CNN filter--property correlation has a direct
antecedent in the Network Dissection
framework~\cite{bau2017netdissect}, which associates
individual filters with semantic concepts by matching
activation maps against a hand-labelled concept dataset.
Our setting is simpler in that the ``concepts'' are
algorithmically computable structural properties (row sums,
adjacent-pair statistics, and so on) rather than
human-annotated semantic categories.  The goal is also
different: we use the correlations to rank which
properties the LLM should reason about during constraint
synthesis, rather than to interpret a vision-trained
network.

\section{Benchmark Context: Hardening and Streamliner Evaluation}\label{sec:hardening}

Before presenting our method we take a page to situate the
evaluation.  Because streamliners do not preserve satisfiability,
a CP practitioner will typically exhaust sound options (standard
symmetry breaking, implied constraints, redundant constraints) before
reaching for streamlining. This makes the choice of baseline
model important for evaluating any automated streamliner
synthesiser.  We report a small empirical observation on the
benchmark suite used by StreamLLM~\cite{voboril2024streamllm}:
on several of their problems the as-shipped baseline model omits
literature-standard hardening, and adding those constraints
substantially changes the instance distribution.

\begin{figure}[!htbp]
\centering
\includegraphics[width=\textwidth]{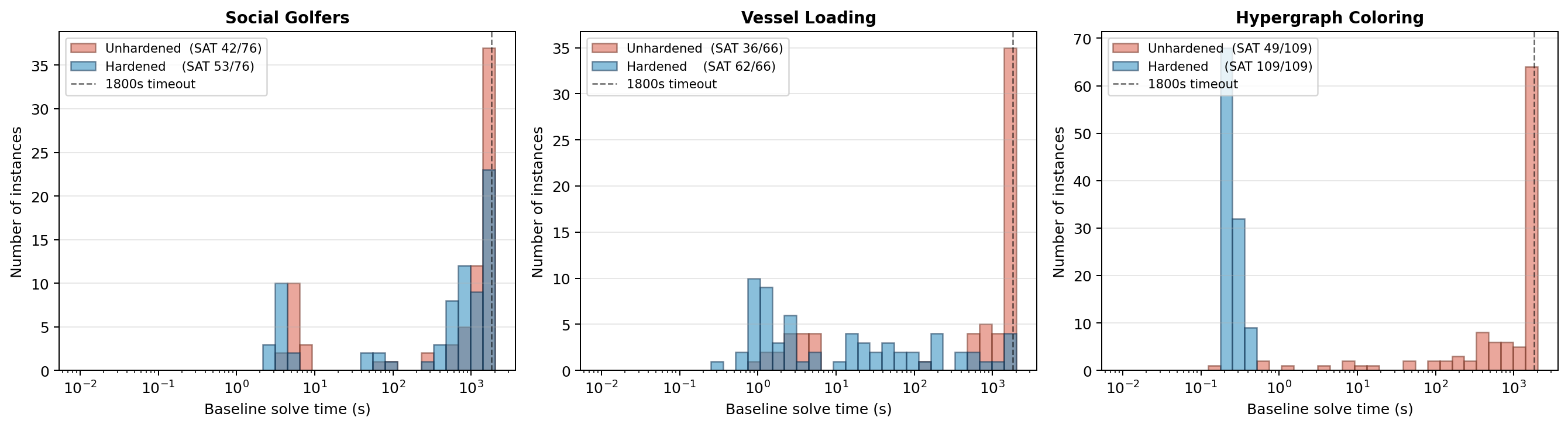}
\caption{Chuffed baseline solve-time distributions for three
StreamLLM benchmark problems before and after literature-standard
hardening (all instances).  Red: unhardened; blue: hardened.
Dashed vertical at \SI{1800}{\second} timeout.}
\label{fig:hardening}
\end{figure}

For three of StreamLLM's benchmark problems we compared the
solve-time distribution of the as-shipped MiniZinc model against
a hardened variant that adds standard constraints from the
literature (for Social Golfers, the week-1 canonical partition
combined with a golfer-1 group anchor and an inter-week lex
ordering~\cite{flener2002rowcolsym}; for Vessel Loading, a lex-leader symmetry break over
identical containers and an orientation canonicalisation for
square containers; for Hypergraph Coloring, value-precedence
on colour labels and the implied vertex-count sum).
Per-problem MiniZinc for these additions appears in
\Cref{app:hardening}.  \Cref{fig:hardening} and
\Cref{tab:hardening-summary} report the result.

\begin{table}[t]
\centering
\caption{Baseline solve-time distribution on StreamLLM test
instances. Chuffed \texttt{0.13.1}, \SI{1800}{\second} timeout.}
\label{tab:hardening-summary}
\begin{tabular}{lrrrrr}
\toprule
Problem & $N$ & \multicolumn{2}{c}{SAT / UNKN} & \multicolumn{2}{c}{Median (s)} \\
\cmidrule(lr){3-4}\cmidrule(lr){5-6}
        &     & Unhardened & Hardened & Unhardened & Hardened \\
\midrule
Hypergraph Coloring  & 94 & 36 / 58 & 94 / \phantom{0}0 & 1800 & \phantom{000}0.2 \\
Vessel Loading       & 51 & 21 / 30 & 48 / \phantom{0}3 & 1800 & \phantom{00}15 \\
Social Golfers       & 61 & 27 / 34 & 38 / 23 & 1670 & 1045 \\
\bottomrule
\end{tabular}
\end{table}

The most striking effect is on Hypergraph Coloring, where
adding the value-precedence chain and the colour-count identity
reduces the median test baseline to \SI{0.2}{\second} and
collapses the timeout count to zero.  Vessel Loading sees a
similar tail compression: the median drops from
\SI{1800}{\second} to \SI{15}{\second} and the unknown count
from \num{30} of \num{51} to \num{3}.  Social Golfers is
meaningfully improved by hardening but remains non-trivial.

The practical implication is that for these problems sound
hardening is not just methodologically prior to streamlining;
it is also where the largest performance gains come from.
Implied and symmetry-breaking constraints are guaranteed to
preserve satisfiability, deliver the order-of-magnitude
baseline compressions reported in
\Cref{tab:hardening-summary}, and are the natural first move
when a CP practitioner faces a slow benchmark model.
Streamliners, which are not solution-preserving, are a
last-resort lever, justified only after standard sound
techniques have been applied.  An automated streamliner
synthesiser evaluated against an unhardened baseline risks
rediscovering constraints a practitioner would have added by
hand before attempting any streamlining at all.  The reported
speedups therefore conflate two distinct effects: the recovery
of missing sound constraints (gains that would disappear once
the model is properly hardened) and genuine beyond-hardening
streamliner discovery.  Only the latter is the intended
contribution of the streamlining literature, and reporting
the two together conflates what streamlining actually
achieves with the recovery of missing model constraints.

This observation is not a criticism of StreamLLM's contribution;
they have been transparent about constraint types throughout.
Rather, it informs our evaluation choice.  For the remainder of
this paper we evaluate our method on hardened variants of
\textbf{Social Golfers} (\Cref{sec:eval-hardened-sg}) and
\textbf{Vessel Loading} (\Cref{sec:eval-hardened-vl}), where
literature-standard symmetry breaking and implied constraints
are already in place and genuine streamliner discovery is the
remaining lever.  We additionally include \textbf{Black Hole}
(\Cref{sec:eval-bh}) as a third evaluation problem.  Black
Hole has no canonical hardening: its sequence is semantically
ordered and the decision variables admit no natural
relabelling group, so neither symmetry-breaking nor
implied-constraint additions are available.  This makes it
the one problem in our suite where a direct head-to-head with
StreamLLM is methodologically clean, as both pipelines must
discover genuine streamliners and neither can recover a
missing model constraint.

\subsection{Composite test set construction}\label{sec:hardening-test}

A practical consequence of evaluating on hardened models is that
many of StreamLLM's distributed test instances become trivial
under hardening, leaving little baseline solve time for any
streamliner to reduce.  We address this by drawing test
instances from two peer-reviewed sources: the
StreamLLM~\cite{voboril2024streamllm} test sets and the
automated-streamliner-portfolios artefact accompanying
Spracklen \emph{et al.}~\cite{spracklen2023portfolios}.  Both
sets were generated by the same informative-instance-generation
framework, AutoIG, of Dang
\emph{et al.}~\cite{dang2022instancegen}.  We retain only test instances on
which the hardened baseline returns SAT within a
\SI{3600}{\second} timeout: the regime where the per-instance
speedup metric is well-defined.  For Social Golfers this yields
\num{61} test instances (\num{52} from~\cite{voboril2024streamllm},
\num{9} from~\cite{spracklen2023portfolios}); for Vessel
Loading, \num{22} test instances (\num{4}
from~\cite{voboril2024streamllm}, \num{18}
from~\cite{spracklen2023portfolios}); for Black Hole, \num{49}
of the \num{53} StreamLLM-distributed instances pass the
SAT-within-\SI{3600}{\second} criterion (the remaining \num{4}
have unknown baselines within the timeout and are excluded).
Black Hole has no canonical hardening, so the BH test set is
drawn entirely from the StreamLLM distribution without
augmentation.

\section{Pipeline}\label{sec:pipeline}

Grammar-based approaches~\cite{spracklen2018montecarlo,spracklen2019optimisation,spracklen2023portfolios}
search a pre-specified constraint lattice: every candidate is
drawn from constraint shapes the designer anticipated, so anything
outside the grammar is unreachable by construction.
StreamLLM~\cite{voboril2024streamllm} removes this ceiling by
prompting an LLM directly with the MiniZinc model text, freeing it
to propose arbitrary constraint shapes.  The cost is that the LLM
is then working from parametric knowledge alone: it knows what
constraints \emph{look like} for this problem class, but has no
evidence of what patterns actually hold in the feasible solutions of
the specific instance at hand.  The obvious alternative (presenting
the LLM with the enumerated solutions directly and asking it to find
patterns) is impractical: even a modest corpus of \num{500}
solutions for a $52 \times 52$ Black Hole instance amounts to tens
of thousands of numbers, and LLMs are poorly suited to reasoning
over raw numerical arrays at that scale, both in terms of context
budget and in terms of the kind of systematic pattern-finding the
task demands, a limitation we encountered directly when testing
uncompressed solution payloads during development.

Our pipeline keeps the LLM's generative freedom while replacing
parametric guessing with grounded evidence.  The key observation is
that a CNN trained to distinguish genuine solutions from perturbed
non-solutions \emph{must} be detecting real structural regularities:
if its filters can reliably separate the two classes, they are
responding to patterns that exist in the solution set rather than
artefacts of the model text.  We extract those filters as explicit
references to solution-level structure and pass them to the LLM,
giving it a concrete signal to reason about rather than asking it
to hypothesise in the dark.  The result is constraint synthesis that
is both unconstrained in shape (the LLM can propose any MiniZinc
expression) and anchored in observed evidence (the proposed
constraint must explain a pattern the CNN demonstrably found).

\begin{figure}[!htbp]
\centering
\begin{tikzpicture}[
  font=\sffamily\footnotesize,
  node distance=4mm and 7mm,
  box/.style={draw, rounded corners=2pt, minimum height=8mm,
              minimum width=22mm, align=center, inner sep=3pt},
  data/.style={box, fill=blue!7, draw=blue!50},
  proc/.style={box, fill=orange!12, draw=orange!60},
  llm/.style={box, fill=green!10, draw=green!50!black},
  val/.style={box, fill=gray!12, draw=gray!60},
  finalnode/.style={box, fill=violet!10, draw=violet!60, minimum width=24mm},
  arrow/.style={->, >=Stealth, semithick, line cap=round},
  pathlbl/.style={font=\sffamily\scriptsize\itshape, inner sep=2pt},
]
\node[data] (instance) {training\\instance};
\node[data, right=of instance] (enum) {solution\\enumeration};
\node[data, right=of enum] (sols) {$N$ feasible\\solutions};

\node[proc, below=8mm of sols, xshift=-22mm] (props) {$\sim$30 structural\\properties};
\node[proc, below=8mm of sols, xshift=22mm] (encode) {tensor\\encoding};

\node[proc, below=8mm of encode] (negs) {perturbed\\negatives};
\node[proc, below=8mm of negs, xshift=-22mm] (cnn) {contrastive\\CNN training};

\node[proc, below=10mm of cnn, xshift=-22mm] (corr) {filter--property\\correlations};
\node[proc, below=10mm of cnn, xshift=22mm] (contrast) {filter-contrast\\solution pairs};

\node[llm, below=8mm of corr] (lstats) {LLM\\(\texttt{llm\_stats})};
\node[llm, below=8mm of contrast] (ldisc) {LLM\\(\texttt{llm\_discovery})};

\node[val, below=10mm of lstats, xshift=22mm] (pool) {candidate\\pool};
\node[val, below=of pool] (cluster) {LLM semantic\\clustering};
\node[val, below=of cluster] (validate) {two-phase\\validation};
\node[finalnode, below=of validate] (final) {validated streamliners};

\draw[arrow] (instance) -- (enum);
\draw[arrow] (enum) -- (sols);
\draw[arrow] (sols) -- (props);
\draw[arrow] (sols) -- (encode);

\draw[arrow] (encode) -- (negs);
\draw[arrow] (negs) -- (cnn);
\draw[arrow] (encode.west) -- ++(-5mm,0) -| (cnn.north);

\draw[arrow] (props) -- (corr);

\draw[arrow] (cnn) -- (corr);
\draw[arrow] (cnn) -- (contrast);

\draw[arrow] (corr) -- (lstats);
\draw[arrow] (contrast) -- (ldisc);

\draw[arrow] (lstats) -- (pool);
\draw[arrow] (ldisc) -- (pool);

\draw[arrow] (pool) -- (cluster);
\draw[arrow] (cluster) -- (validate);
\draw[arrow] (validate) -- (final);

\end{tikzpicture}
\caption{End-to-end pipeline from a MiniZinc model and training
instance to a pool of validated streamliners.  Stages are
colour-coded: \emph{blue} = solution preparation, \emph{orange}
= CNN pattern recognition, \emph{green} = LLM constraint
synthesis, \emph{grey} = validation.  Solutions are enumerated
and encoded into tensors; a CNN is trained contrastively against
perturbed negatives and emits two complementary signals:
per-filter property correlations and high/low filter-activation
solution pairs.  Two LLM-prompting paths consume these signals:
\texttt{llm\_stats} reasons over property statistics with the
correlation prior, and \texttt{llm\_discovery} hypothesises
constraints from the contrast pairs.  Candidates from both
paths are pooled across training instances and ensemble seeds,
semantically clustered, and validated in two phases against
cached baselines.}
\label{fig:pipeline}
\end{figure}
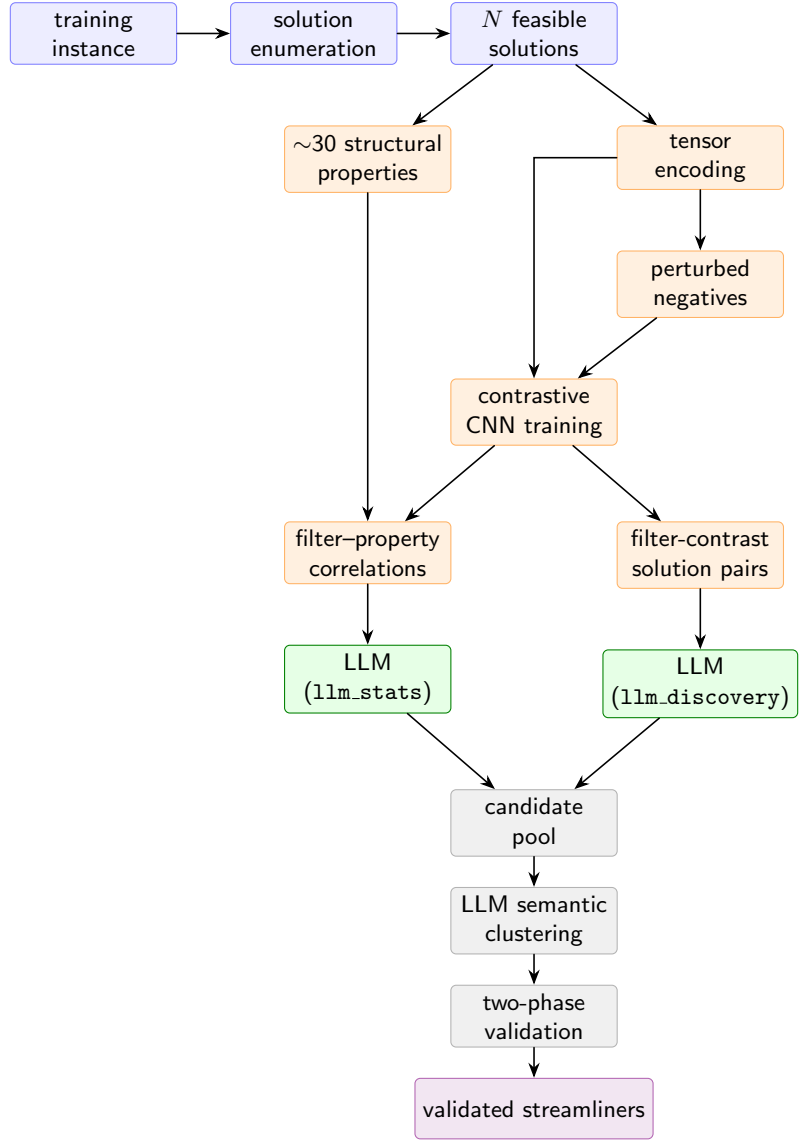

\Cref{fig:pipeline} sketches the end-to-end flow. The pipeline
factors into four stages: solution enumeration, contrastive CNN
training, LLM-driven constraint synthesis along two prompting
paths, and multi-instance pooling with validation.

\subsection{Solution enumeration and encoding}\label{sec:pipeline-enc}

Given a MiniZinc model and a training instance, we enumerate
at least $N = \num{500}$ feasible solutions using Chuffed~\cite{chu2018chuffed}
with the \texttt{all\_solutions} flag (more when the per-instance
enumeration budget allows; some training instances supply
several thousand solutions).  Each solution is encoded as a 2D or 3D
tensor whose shape matches the decision variable's structure:
scalar grayscale for matrix problems, permutation matrix for
\mbox{1-D} permutations, one-hot encoding for assignment-style
problems, and a custom \emph{container rasterisation} for
multi-variable packing problems such as Vessel Loading (where the
decision variables encode container coordinates rather than a
single matrix).  This tensor representation is what the CNN
operates on; the original MiniZinc decision variable is also
retained so the LLM can be shown raw solution samples in their
original shape.

We additionally compute a problem-shape-specific library of
$\approx 30$ structural properties per solution: row sums, column
sums, diagonal sums, value ranges, monotonicity indicators, adjacent
position differences, positional centroids, extremum locations, and
similar invariants.  The exact set is conditional on the variable
shape, but each property is a real-valued summary of one solution.
A property whose value is constant across all enumerated solutions
is an \emph{implied} consequence of the original constraints (never
useful as a streamliner).  A property whose value lies in a narrow
band across solutions, but is not strictly constant, is a
first-class streamliner candidate: tight bounds near the observed
range often retain satisfiability while pruning aggressively.

\paragraph{Problem applicability.}
The pipeline applies to any CSP whose decision variables can be
encoded as a fixed-shape 2D or 3D tensor.  Matrix problems
(Latin squares, BIBD, magic squares) map naturally to grayscale
images; permutation problems (N-Queens, Black Hole) to binary
permutation matrices; assignment problems to one-hot encodings;
and packing problems with coordinate variables (Vessel Loading)
to rasterised spatial maps.  The key requirement is that the
solution structure is spatially regular enough for a
convolutional filter to detect recurring patterns: problems
whose solutions have highly variable structure across instances,
or whose decision variables do not admit a natural 2D layout,
are less suitable candidates.  A concrete negative example is
Steiner Triple Systems, where every feasible solution has
identical row sums, column sums, and pair counts by
construction; the property library finds nothing with
non-zero variance and the \texttt{llm\_stats} path produces no
useful candidates (\Cref{sec:threats}).  Conversely, graph
colouring, scheduling, and routing problems with a fixed
underlying topology are natural future targets, provided the
encoding maps the problem's structure to a spatially
interpretable representation (e.g.\ adjacency matrix or
timeline grid) before the CNN stage.

\subsection{Contrastive CNN training and pattern recognition}\label{sec:pipeline-cnn}

Per-instance, we train a 3-layer Convolutional Neural Network
($32{\to}64{\to}128$ channels, $3{\times}3$ kernels, batch
normalisation, ReLU) contrastively on enumerated solutions versus
\emph{negative samples}: tensors generated by perturbing a feasible
solution into a structurally close but typically infeasible
configuration.  We use three negative generators per solution
shape (row-permuted, position-swapped, and fully uniform-random)
and balance positives and negatives one-to-one.  The
classification objective forces the CNN to learn filters that
detect structural patterns the feasibility predicate exploits:
filters that activate on real solutions and not on random
arrangements of the same domain values.

The CNN's signal is surfaced in two complementary ways for
downstream constraint synthesis.  Filter selection operates
across all three conv layers: for each layer we rank filters
by per-solution mean-activation variance and retain the top
six per layer (eighteen filters in total; figures in this
paper show the top three for visual clarity).  This keeps both
shallow filters that detect local primitives (e.g.\ adjacency
patterns) and deeper filters that compose those primitives
into longer-range structure; both are useful inputs for
the LLM's constraint synthesis, and pre-committing to one
depth would discard signal.  In practice, however, the most
spatially interpretable constraint-discovery signal in our
evaluation was concentrated in the first convolutional layer:
the deeper layers showed broadly similar spatial patterns
without clear evidence of qualitatively distinct hierarchical
features.  Whether multi-layer extraction adds meaningful
constraint-discovery value over a shallower architecture
remains an open question (\Cref{sec:threats}).

First, we compute Pearson
correlations between each retained filter's per-solution
activation vector and each structural property's per-solution
value vector.  Properties with high $|r|$ correlations are
properties the CNN's discriminative behaviour explains, and
hence first-class targets for streamlining.  Second, we
identify high-activation and low-activation solution pairs
under each retained filter; these contrast pairs feed the
filter-contrast LLM path described below.

We use an ensemble of $K=3$ random initialisation seeds per
training instance to reduce filter-selection variance: different
CNN initialisations learn slightly different filter bases, and
pooling candidates across the ensemble yields a more robust
relevance ranking than any single seed.

\begin{figure}[!htbp]
\centering
\includegraphics[width=\textwidth]{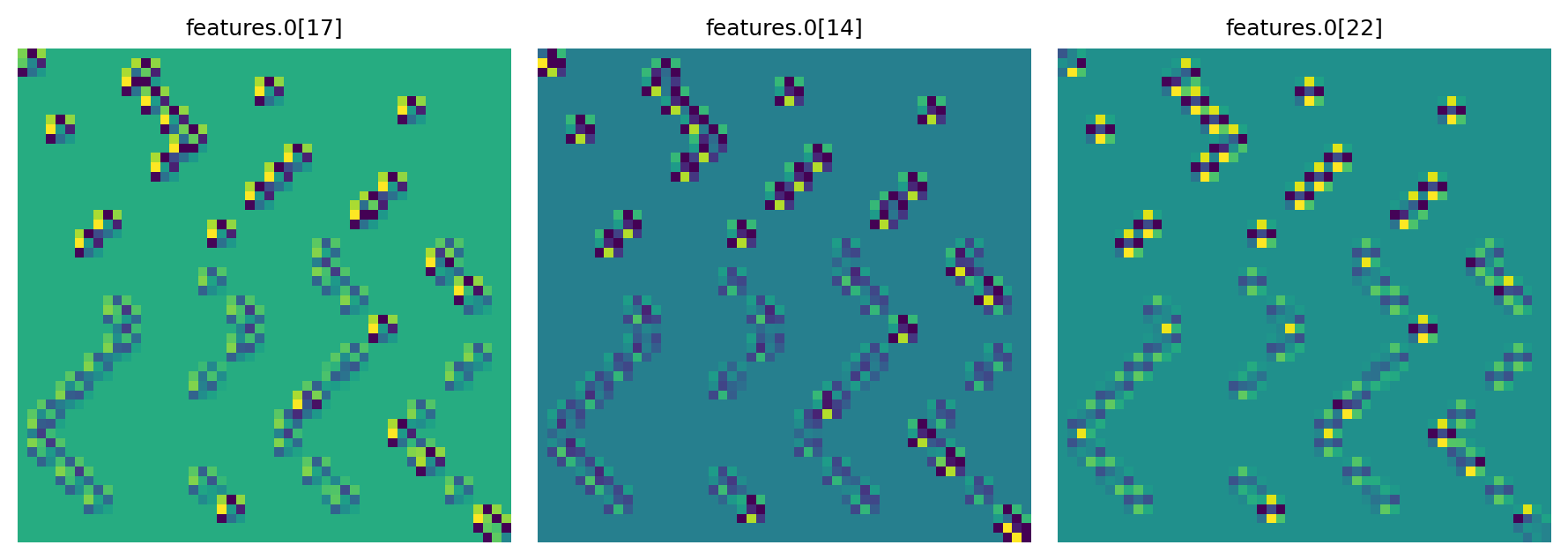}
\caption{Mean activation maps for the three top-variance
\texttt{features.0} filters of a CNN trained contrastively on
Black Hole solutions versus near-miss negatives (training
instance \texttt{1191}; mean over the enumerated-solutions
corpus).  Each filter exhibits a zigzag/staircase signature: the
rank-adjacency constraint forces consecutive sequence positions
to play cards of adjacent rank, so the 1s in neighbouring rows
of the permutation matrix shift by a small column offset as
the rank walk ascends or descends, tracing a diagonal staircase
across the canvas.  The CNN learns this dominant rank-adjacency structure
without being told what to look for; this signal is what
grounds the LLM constraint synthesis in
\Cref{sec:pipeline-llm}.  Deeper layers
(\texttt{features.3}, \texttt{features.6}) track the same
primitive at longer composition lengths.}
\label{fig:cnn-activations-bh}
\end{figure}

\Cref{fig:cnn-activations-bh} renders this learned structure on
a Black Hole training instance: every top-variance filter at
layer 0 encodes the rank-adjacency rule that defines the
problem.

\subsection{LLM-driven constraint synthesis}\label{sec:pipeline-llm}

The CNN's pattern-recognition output and the per-property
statistics are translated into candidate MiniZinc streamliners
by a Large Language Model along two prompting paths.

\paragraph{Statistics path (\texttt{llm\_stats}).}
We assemble a JSON payload containing (i) per-property summary
statistics (mean, standard deviation, range, near-constant flag,
median); (ii) the top three CNN filters by absolute correlation
for each property, ranked; (iii) up to five raw sample solutions
in their original variable shape; and (iv) when training on
multiple instances of varying size, a per-property progression
table that allows the LLM to extrapolate bounds parametrically in
$n$ rather than pinning to a single observation.  The prompt
instructs the LLM to emit a JSON array of candidate MiniZinc
constraints, each tagged with the originating property, an
\emph{aggressiveness} level (\emph{conservative} = bound has a
safety margin past the observed extreme; \emph{tight-fit} = bound
matches the observed extreme; \emph{aggressive} = bound is the
LLM's parametric extrapolation, possibly tighter than any
observation), and a syntactic form (existential, universal,
aggregate, pairwise).  We sample multiple
(form, aggregation, aggressiveness) combinations per call to
avoid the LLM collapsing into a single shape.

\paragraph{Filter-contrast path (\texttt{llm\_discovery}).}
For each top-variance CNN filter we partition the solution corpus
into a high-activation group and a low-activation group, pass both
as labelled numerical data to the LLM, and ask it to
hypothesise what structural pattern distinguishes the two groups.
The prompt asks for a textual hypothesis followed by a candidate
MiniZinc constraint that encodes the hypothesised pattern.  Unlike
the statistics path, this path does not anchor the LLM to a
predefined property library: the LLM is free to propose constraint
shapes outside the property catalogue, including
multi-variable structural relationships and non-trivial
canonicalisations that the property statistics alone do not
suggest.

\paragraph{Mechanical templates (briefly).}
For comparison we also implemented a deterministic template path:
near-constant properties at observed value $k$ instantiate
\verb|constraint p = k| directly, with similar mechanical
substitutions for monotonicity flags and extremum-shaped
properties.  Templates are reproducible and cost-free, but in our
hardened-model evaluation they consistently underperform both LLM
paths, which generalise to constraint shapes the templates do not
encode.  We retain templates as a baseline-stage in the pipeline
but report them only when relevant.

All three sources (statistics path, filter-contrast path,
and templates) emit method-tagged candidates that are
text-deduplicated on exact constraint string.  Templates are
preserved as a baseline column in the validation pool but on
the hardened-evaluation regime contribute no headline picks
(\Cref{fig:pareto-front}).

\subsection{Multi-instance pooling and validation}\label{sec:pipeline-valid}

The previous stages run independently on each of $M$ training
instances.  We pool all method-tagged candidates across instances
and seeds, and a second LLM call performs semantic
clustering: candidates whose constraint shapes match modulo
parameter substitution are grouped, and from each cluster we
expand up to four representatives (tightest, loosest, parametric
median, and an extrapolated variant) into the validation
candidate set.  Semantic clustering is necessary because text
deduplication misses parametrically-equivalent constraints that
arise across different instance sizes.

Each candidate is evaluated on the training instances to filter
to those that improve over baseline somewhere and to produce the
training-set savings score used for portfolio selection
(\Cref{sec:pipeline-portfolio}).  Each solve is capped at the
baseline time on the same instance (a streamliner slower than
baseline is not useful) and per-pair results are incrementally
checkpointed so that the pipeline is resumable.  The final
reported metrics (\Cref{sec:eval}) are computed on the held-out
test instances described in \Cref{sec:hardening-test}; no
separate validation set is used between training and test.

\subsection{Portfolio deployment via family-budget allocation}\label{sec:pipeline-portfolio}

The validated candidate pool is the input to a portfolio
selection step that produces the deployed streamliner set.
The natural baseline rule (\emph{simple top-$k$}: rank
candidates by training-set savings and pick the top-$k$)
is blind to a structural property of the LLM-driven pool:
constraint variants of the same shape (different
aggressiveness levels, different parameter regimes) typically
all rank highly because they all save substantial training
time, but contribute redundantly to the deployed portfolio.
On Black Hole, the simple top-3 picks are all variants of the
pile-top $y$-coordinate bound; the deployed compute budget is
dominated by one constraint family even though the pool
contains other useful shapes.  We diagnose this in
\Cref{sec:analysis-selection}.

We deploy a \emph{family-budget portfolio allocation} that
explicitly enforces both inter-family and intra-family
diversity.  The mechanism repurposes the LLM's own constraint
descriptors (short snake-case labels emitted alongside each
synthesised constraint: \texttt{pile\_top\_early\_25},
\texttt{max\_adj\_diff\_aggressive},
\texttt{rank\_walk\_endpoints}, etc.) as semantic family
tags;
candidates are bucketed by descriptor prefix.  For a
deployment of $k$ families with $m$ members per family:
\begin{enumerate}
\item Select the top $k$ families ranked by the highest
individual training-set savings of any member of that family.
\item Within each selected family, retain the top $m$ members
by individual training-set savings.
\item Race all $k \cdot m$ retained streamliners against the
baseline.  Each family receives a wall-clock budget equal to
the baseline time; within a family, the $m$ retained members
share that budget equally (slot $= t_{\rm base} / m$).  A
streamliner ``succeeds'' for its family if it returns SAT
within its slot.  The family's contribution is the
wall-clock time at which its first member returns SAT (not
the individual streamliner's solve time), or the baseline
time if no member succeeds.
Portfolio time per instance is the minimum of the $k$ family
contributions and the baseline.
\end{enumerate}
Total compute is identical to a simple top-$k$ $(k+1)$-way
race: $k$ family-races plus the baseline solve, with each
family-race using $t_{\rm base}$ of compute distributed across
$m$ slots.  Critically, each individual streamliner receives
\emph{less} wall-clock time than in a simple top-$k$ race
(slot $= t_{\rm base}/m$ rather than $t_{\rm base}$); the
family-budget allocation trades per-streamliner time for
coverage across aggressiveness levels, not additional compute.  Two complementary diversity properties motivate
this allocation.  \emph{Inter-family} diversity (controlled by
$k$) ensures the portfolio covers structurally distinct
constraint shapes.  \emph{Intra-family} diversity (controlled
by $m$) covers different aggressiveness levels and parameter
regimes within each shape.  We expect diminishing returns in
$m$ because variants within a family target the same property
and share retention behaviour at the margin; we observe this
empirically (\Cref{sec:analysis-family-sensitivity}, with the
sweet spot at $m = 3$).

Throughout the evaluation we deploy family-budget allocation at
$k = 3$, $m = 3$, yielding a portfolio of \num{9} streamliners
across \num{3} families per problem.
We additionally report simple top-3 numbers as a comparison
baseline; for direct comparison with prior work
\cite{voboril2024streamllm} on Black Hole we keep the
simple top-3 reading as well.

The mechanism is independent of the LLM that generated the
constraints: any constraint-synthesis pipeline whose pool
members carry semantic labels (or that admits a downstream
clustering pass) can adopt the same allocation.  In our case
the labels are free (the LLM emits them as part of each
constraint's description during synthesis,
\Cref{sec:pipeline-llm}), so family-budget allocation adds
no incremental cost to the pipeline.

\section{Evaluation}\label{sec:eval}

\subsection{Experimental setup}\label{sec:eval-setup}

\paragraph{Solver and protocol.}
We use the Chuffed \texttt{0.13.1}
solver~\cite{chu2018chuffed} under MiniZinc
\texttt{2.9.5}~\cite{nethercote2007minizinc}, with a
\SI{3600}{\second} wall-clock timeout and a fixed random seed
(\num{42}) for every solve.  Baselines are computed once per
test instance and cached; each streamlined solve is individually
capped at the cached baseline time on the same instance (a
streamliner that runs longer than the baseline is by construction
unhelpful).  Our CNN is trained for \num{100} epochs per training
instance with an ensemble of $K=3$ random seeds.  Constraint synthesis uses Anthropic's Claude Opus models.
We additionally tested locally hosted open-weight models via
Ollama (\texttt{qwen2.5-coder:14b}, the largest model our
hardware could run), but the generated MiniZinc was not
reliable enough to sustain the synthesis pass; the open-weight
limitation is discussed further in \Cref{sec:threats}.

\paragraph{Hardware.}
All experiments ran on a single workstation: an Intel Core
i7-13700K CPU (\num{16} cores, \num{24} threads, up to
\SI{5.4}{\giga\hertz}), \SI{32}{\giga\byte} of system memory,
an NVIDIA GeForce RTX 4090 GPU (\SI{24}{\giga\byte} VRAM), and
a \SI{1}{\tera\byte} SSD.  The host OS is Ubuntu \num{22.04} on
WSL2 (Linux kernel \num{5.15}) and the full pipeline executes
inside Docker (Python \texttt{3.11}, PyTorch, MiniZinc \texttt{2.9.5}
with the Chuffed solver).  CNN training uses the GPU; baseline
and streamlined solves use Chuffed on a single CPU thread per
solver process, with parallel portfolio races scheduled across
separate processes.  Each solver process runs under
\texttt{runsolver}~\cite{roussel2011runsolver}, which enforces per-solve
wall-clock and memory limits and returns a clean timeout signal
rather than relying on solver-internal termination.

\paragraph{Test sets.}
For Black Hole we use the StreamLLM-distributed test
instances~\cite{voboril2024streamllm} unmodified
(\Cref{sec:hardening} explains why no canonical hardening
applies to BH).  For hardened Social Golfers and
Vessel Loading we use the composite test set described in
\Cref{sec:hardening}: the union of the StreamLLM-distributed
instances and the automated-streamliner-portfolios
artefact~\cite{spracklen2023portfolios}, restricted to instances
on which the hardened baseline returns SAT within the
\SI{3600}{\second} timeout.  This yields \num{61} instances for
hardened Social Golfers (\num{52} from~\cite{voboril2024streamllm},
\num{9} from~\cite{spracklen2023portfolios}), \num{22} for
hardened Vessel Loading (\num{4} and \num{18} respectively), and
\num{49} of \num{53} StreamLLM Black Hole instances (the
remaining \num{4} have unknown baselines within
\SI{3600}{\second} and are excluded by the same SAT-within-
timeout criterion applied to the SG and VL test sets).

\paragraph{Metrics.}
For each test instance $i$ let $t_b(i)$ be the cached baseline
solve time and $t_c(i)$ the streamlined solve time of candidate
$c$.  The per-instance speedup is $t_b(i)/t_c(i)$ when both
return SAT.  Aggregated metrics reported below:
\begin{itemize}
\item \emph{Portfolio time savings.} We race the deployed
portfolio (either simple top-3 or family-budget) against the
baseline in parallel and report
$(\sum t_b(i) - \sum t_{\text{winner}}(i))/\sum t_b(i)$,
where $t_{\text{winner}}(i)$ is the wall-clock time of the
first process (streamliner or baseline) to return SAT on
instance $i$.  For simple top-3 this is a 4-way race;
for family-budget at $k=3$, $m=3$ it is a 10-way race within
the same total compute envelope.  This mirrors the metric
reported in Figure~3 of~\cite{voboril2024streamllm},
converted from their reduction form $1 - t_{\rm new}/t_{\rm old}$.
\item \emph{Pool ceiling.} At each instance we pick the single
fastest SAT-returning streamliner across the \emph{entire}
validated pool (all candidates, not just the deployed subset),
then aggregate.  This is an oracle upper bound: no runtime
selection scheme drawing from the same pool can exceed it.
The gap between a deployed portfolio (simple top-3 or
family-budget, both of which select a small subset of the pool)
and the ceiling quantifies how much selection inefficiency
leaves on the table.
\item \emph{Best-single geometric-mean speedup.} For the single
top-performing streamliner per problem, the geometric mean of
per-instance speedups over instances where both the streamliner
and the baseline returned SAT.  This complements the portfolio
metric: a streamliner with high per-instance speedup but narrow
SAT-retention is rewarded here even if it does not survive
portfolio aggregation.
\item \emph{CPU-adjusted time savings.}  The $(k+1)$-way race
consumes up to $(k+1)$ times the CPU of a lone baseline solve,
since all parallel processes run until the winner finishes.  We
therefore also report \emph{CPU-adjusted savings}
$\bigl(\sum t_b(i) - (k+1)\sum t_{\rm winner}(i)\bigr)/\sum t_b(i)$
as an honest accounting of total compute cost.  We regard the
wall-clock figure as the primary practical metric: modern
hardware routinely provides the spare cores needed to run a
small parallel portfolio at no wall-clock penalty, and this is
the deployment model assumed by StreamLLM~\cite{voboril2024streamllm};
Spracklen \emph{et al.}~\cite{spracklen2023portfolios} account
for CPU cost explicitly in their portfolio evaluation.  The
CPU-adjusted figure is reported for completeness and for
practitioners operating under strict compute budgets.
\end{itemize}

\paragraph{Selection rule.}
Throughout the evaluation we deploy the family-budget
allocation introduced in \Cref{sec:pipeline-portfolio} at
$k = 3$ families and $m = 3$ members per family, racing
\num{9} streamliners against the baseline within the same
compute envelope as a 4-way race.  As a comparison baseline
we also report the simple top-3 rule (rank pool members by
training-set savings; deploy the top 3): this is the natural
single-streamliner-per-slot reading and the closest analogue
to prior portfolio work.  Family-budget is our default
deployment; simple top-3 is reported alongside for direct
comparison with prior published portfolios.

\subsection{Overview: speedup heatmap}\label{sec:eval-overview}

\begin{figure}[!htbp]
\centering
\includegraphics[width=\textwidth]{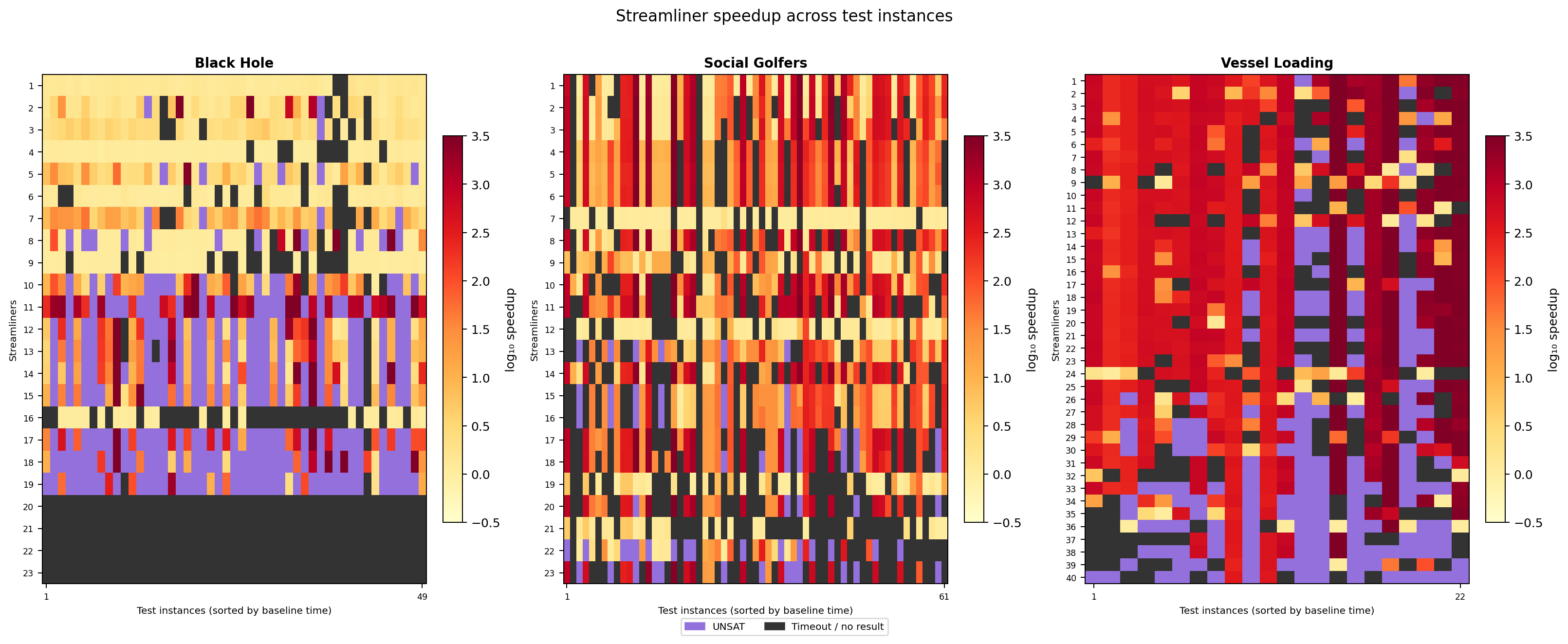}
\caption{Speedup heatmap across all validated streamliners and
test instances for Black Hole (\num{23} streamliners,
\num{49} instances), Social Golfers (\num{23} streamliners,
\num{61} instances), and Vessel Loading (\num{40} streamliners,
\num{22} instances).  Rows are ordered by SAT-retention count
(highest coverage first); columns by baseline solve time
(easiest left).  Colour encodes $\log_{10}$ speedup on
SAT-retained instances (yellow $\approx 1\times$, dark red
$\geq 1000\times$); purple = UNSAT; dark grey = timeout or
no result.  The BH block rows~\num{20}--\num{23}
(all dark grey) are the template candidates, which contribute
nothing on this evaluation regime.}
\label{fig:speedup-heatmap}
\end{figure}

\Cref{fig:speedup-heatmap} gives a bird's-eye view of the full
validated pool across all three problems before the per-problem
analysis.  Each row is one validated streamliner, ordered by
SAT-retention count; each column is one test instance, ordered
by baseline solve time.  Colour encodes $\log_{10}$ speedup
where the streamliner returns SAT; purple marks instances
rendered unsatisfiable; dark grey marks timeout or no result.

\Cref{tab:results-combined} summarises the headline portfolio
metrics across all three problems.

\begin{table}[t]
\centering
\caption{Portfolio time savings across all three problems.
\emph{Pool ceiling} is an oracle upper bound: at each instance,
the single fastest SAT-returning streamliner from the full
validated pool.  \emph{FB$\to$ceiling gap} is the residual
between family-budget and the pool ceiling.  StreamLLM
comparison is only reported for Black Hole, the one problem
where both pipelines face the same (unhardened) baseline.}
\label{tab:results-combined}
\begin{tabular}{lrrr}
\toprule
 & \textbf{Black Hole} & \textbf{Hardened SG} & \textbf{Hardened VL} \\
 & (49 instances) & (61 instances) & (22 instances) \\
\midrule
Simple top-3                   & \num{79.7}\% & \num{94.4}\% & \num{97.4}\% \\
Family-budget ($k=3$, $m=3$)   & \textbf{\num{89.4}\%} & \textbf{\num{98.6}\%} & \textbf{\num{98.8}\%} \\
Pool ceiling                   & \num{94.9}\% & \num{98.7}\% & \num{99.6}\% \\
FB$\to$ceiling gap             & 5.5\,pp      & 0.1\,pp      & 0.8\,pp \\
\midrule
StreamLLM (realtime, Fig.~3)   & $\approx$\num{80}\% & --- & --- \\
\bottomrule
\end{tabular}
\end{table}

\subsection{Black Hole}\label{sec:eval-bh}

Black Hole~\cite{gent2007blackhole,gent1999csplib} is a
single-player card game encoded as a CSP: arrange
\num{51} of \num{52} playing cards into a sequence and a layout
of \num{17} piles of \num{3} cards, such that each consecutive
pair of cards in the sequence differs by $\pm 1$ modulo \num{13}
(rank adjacency) and a small set of layout constraints relating
piles to the sequence are satisfied.  The key decision variables
are \texttt{x[p]}, the card at position~$p$ in the play
sequence, and \texttt{y[c]}, its inverse (the position of
card~$c$ in the sequence).  The parameter \texttt{layout[i,j]}
gives the card in pile~$i$ at depth~$j$, so
\texttt{layout[i,1]} is the initial top card of pile~$i$,
the only card in that pile that is accessible at the start of
the game.

\paragraph{Pipeline output.}
Training on the \num{16} StreamLLM training instances, our
pipeline generates a pool of \num{23} candidate streamliners
(\num{6} from the LLM-statistics path, \num{4} from the template
path, \num{13} from the LLM-discovery path).  The simple-top-3
ranking (training-savings only) selects three pool members all
from the LLM-discovery \emph{pile-top} family (we use
\emph{king-pin} as shorthand for the terminal-card constraint
\verb|(x[52]-1) mod 13 = 12|, which requires the last card
played to be a King):
\begin{enumerate}
  \item \verb|forall(i in 1..17)(y[layout[i,1]] <= 25)|
  \item \verb|forall(i in 1..17)(y[layout[i,1]] <= 35)|
  \item \verb|forall(i in 1..17)(y[layout[i,1]] <= 30)| $+$ king-pin
\end{enumerate}
Here \texttt{layout[i,1]} is the \emph{initial} top card of
pile~$i$ (before the game begins) and \texttt{y[c]} is the
position of card~$c$ in the play sequence.  The model already
enforces that a buried card cannot be played before the card
on top of it; the pile-top constraint says something stronger
and non-obvious: the \num{17} initially-accessible cards do
not merely precede their pile-mates, they cluster in the first
$K$ positions of the \emph{global} \num{52}-card sequence.  A
valid solution could in principle place a pile-top card at
move~\num{48} and still satisfy rank-adjacency; the CNN
observed across \num{2000} enumerated solutions that this
almost never happens.  The filter-contrast path surfaced this pattern; the statistics
path missed it as the property library contains no
$y$-coordinate aggregates over the layout indexing array.
This constraint is unreachable by any approach anchored to a
fixed property library, and the raw-solutions alternative is
impractical at this data scale: exhaustively computing variance
over all variable-index combinations in a $52 \times 52$
encoding yields an unranked flood of near-constants with no
signal about which are discriminative.  The CNN finds the
pile-top pattern without being told to look for layout
coordinates: the contrastive objective attends to pile-top
positions because they are the most discriminative signal
separating real solutions from perturbed near-misses.

\paragraph{The discovery mechanism in action.}
The \texttt{llm\_discovery} path operates in two complementary
modes depending on how the CNN signal manifests.  In the
\emph{variance-driven} mode, a filter tracks a per-solution
scalar property with non-trivial variance across the corpus:
for example, filter \texttt{features.6[75]} on training
instance \texttt{1191} correlates strongly with the
\emph{ascending-pairs count} ($r = +0.57$ across \num{2000}
solutions), which counts the number of positions $p$ where
$x[p+1] > x[p]$ (i.e.\ the card value increases from one
sequence position to the next).  The \num{3} highest-activation solutions
under this filter have ascending-pairs counts of
$\{29, 30, 29\}$ out of \num{51} adjacent positions;
the \num{3} lowest have $\{24, 26, 24\}$.  The LLM receives
these contrasting solution groups as numerical data and
converts the observed separation into a bound constraint
(\verb|sum(p in 1..51)(bool2int(x[p+1] > x[p])) <= 27|).
This filter is a non-deployed pool member.  On the test set
the constraint retains SAT on \num{41} of \num{49} instances
and improves on all of them, but the geomean speedup is only
\num{1.1}$\times$: bounding the ascending-pairs count to
$\leq 27$ prunes very little of the search space because most
valid solutions naturally sit near that value (population mean
$\approx 27.2$, std $\approx 1.5$).  The pile-top family
dominates the deployment ranking by a wide margin.

\begin{figure}[!htbp]
\centering
\includegraphics[width=\textwidth]{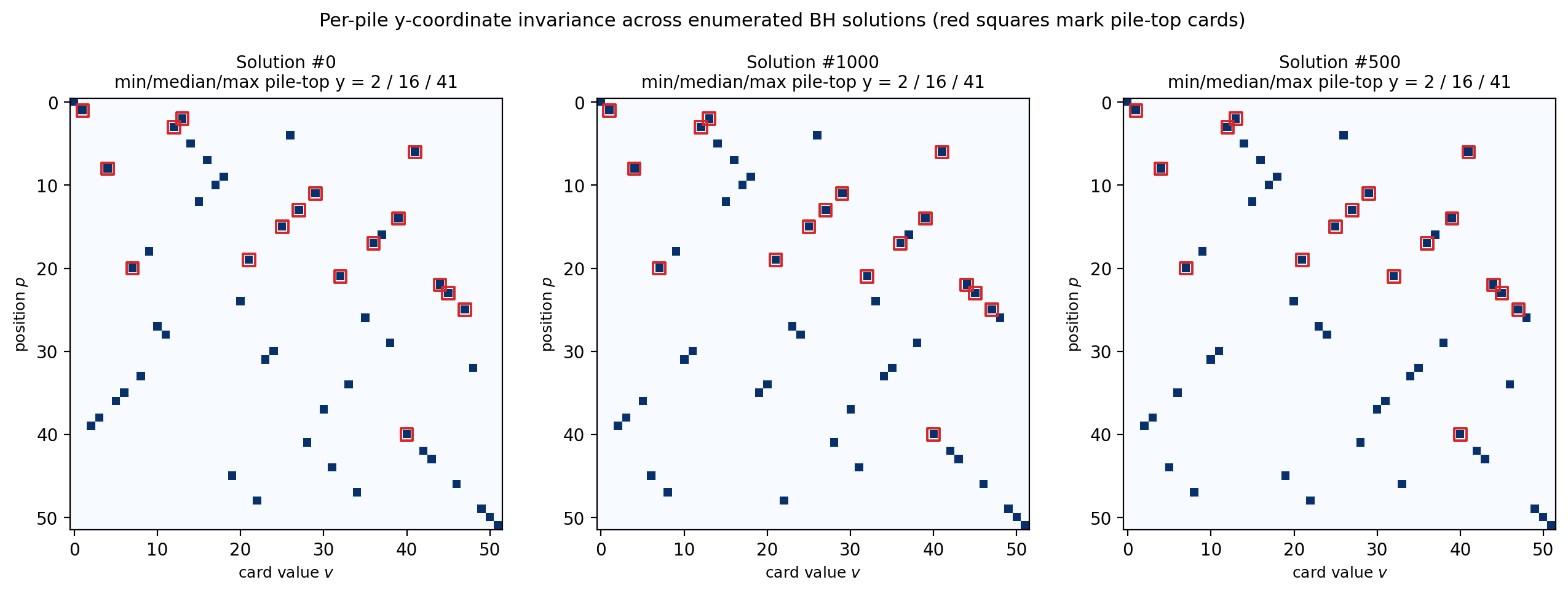}
\caption{Per-pile $y$-coordinate invariance on Black Hole
training instance \texttt{1191}.  Three arbitrary solutions
(positions \num{0}, \num{500}, \num{1000} in the
enumerated corpus) rendered as $52 \times 52$ permutation
matrices.  Red squares mark the \num{17} pile-top cards
(\texttt{layout[i,1]} for $i \in 1..17$).  The red-marked
cells appear at identical $(p, v)$ co-ordinates in all three
panels, while the unmarked cells (non-pile-top cards)
visibly migrate; the invariance is structural, not
accidental.  This figure shows one example of the pattern
the CNN exposes: for this deal, pile-top positions are
fixed across solutions, suggesting a bound on how early
they appear in the sequence.  Pinning pile-tops to exact
positions would not generalise across instances, since the
deal changes per instance and different cards sit on top of
the piles; a parametric bound \texttt{y[layout[i,1]] <= K}
captures the structural regularity without depending on
instance-specific card identities.}
\label{fig:bh-pile-top-invariant}
\end{figure}

In the \emph{invariant-driven} mode, the CNN exposes a
per-element pattern that holds almost universally across
the corpus.  Per-pile $y$-coordinates are nearly invariant
across the \num{2000} enumerated solutions of instance
\texttt{1191}: \num{15} of \num{17} pile-top positions
take a single value, and the remaining two vary by $\pm 1$.
\Cref{fig:bh-pile-top-invariant} makes this directly
visible: in three arbitrary solutions sampled from the
corpus, the same \num{17} pile-top cards land at the same
$(p, v)$ co-ordinates while the non-pile-top cards visibly
migrate.  The LLM could read the maximum pile-top position
(\num{41} on this instance) from the solution data and emit
\verb|y[layout[i,1]] <= K| at three aggressiveness levels;
it is this family that dominates the deployed portfolio.
Both modes feed the same prompt; on Black Hole only the
invariant-driven branch produces deployed picks.

\paragraph{Family-budget portfolio.}
The simple top-3 ranking concentrates the deployed compute
budget on a single constraint family.  The family-budget
allocation (\Cref{sec:pipeline-portfolio}) at $k = 3$ instead
selects three different families (\emph{pile-top},
\emph{max-adj-diff}, \emph{rank-walk-endpoints}) and retains
the top $m = 3$ members of each (\num{9} streamliners total),
racing them within the same compute envelope.  \emph{Pile-top}
contributes its three aggressiveness-level variants;
\emph{max-adj-diff} contributes a tight-fit and an aggressive
variant; \emph{rank-walk} has only one member at this pool size.

\paragraph{Results.}
\Cref{tab:results-combined} reports the headline portfolio metric.  The
deployed family-budget portfolio achieves
\num{89.4}\% time savings, comfortably above StreamLLM's
realtime configuration ($\approx$\num{80}\% from their
Figure~3) and recovering most of the way to the \num{94.9}\%
pool ceiling.  The simple-top-3 reading reaches \num{79.7}\%,
closely matching StreamLLM's published number for like-for-like
comparison, but is structurally limited by the within-pile-top
redundancy of its three picks (diagnosed in
\Cref{sec:analysis-selection}).  The \num{9.7}-percentage-point
gap between simple-top-3 and family-budget is the structural
improvement the family-aware allocation buys.

The qualitative difference between the two pipelines' discoveries
is striking.  StreamLLM's realtime BH triple
(\texttt{x[26]=26}, a rank-descent-forbidding forall, and
\texttt{x[52]=52}) pins individual cards to individual positions,
constraints plausible from reading the model text alone.
Our deployed family operates on \texttt{y[layout[i,1]]}, a
compound expression that composes the layout parameter with the
positional variable to constrain all \num{17} pile-top cards
to appear within the first $K$ positions of the play sequence.  This cross-variable relationship is not visible
in the model text; it emerges from observing that pile-top
positions are nearly invariant across the enumerated solution
corpus.  The contrast illustrates the complementary roles of
the two approaches: model-text-grounded synthesis recovers
plausible structural guesses, while solution-grounded synthesis
discovers structural regularities that only become visible in
the solution distribution.

The best single streamliner,
\verb|forall(i in 1..17)(y[layout[i,1]] <= 25)|, at the most
aggressive end of the pile-top family --- achieves a
geometric-mean speedup of \num{1103}$\times$ over the
\num{26} of \num{49} instances on which it retains
satisfiability, with a maximum per-instance speedup of
\num{4547}$\times$.  \Cref{tab:bh-pile-top} reports all three
deployed pile-top variants and makes the aggressiveness--coverage
tradeoff concrete.  Tightening the bound from $K = 35$ to
$K = 25$ reduces SAT retention from \num{39} to \num{26}
instances but lifts the geometric-mean speedup from
\num{10}$\times$ to \num{1103}$\times$: the constraint prunes
so aggressively that where it applies the solver barely
searches.  Soundness is instance-specific: whether
$y[\texttt{layout}[i,1]] \le K$ holds depends on the particular
Black Hole card layout, not on how hard the instance is to
solve, so baseline difficulty is not a predictor of SAT
retention.  The family-budget portfolio races all three variants
within the pile-top slot, covering the instances that the most
aggressive member fails.

\begin{table}[t]
\centering
\caption{Pile-top family: aggressiveness--coverage tradeoff on the
\num{49} Black Hole test instances.  SAT = instances on which the
streamliner returns SAT within the baseline cap; UNSAT = instances
rendered unsatisfiable.  Geomean and max speedup computed over SAT
instances only.}
\label{tab:bh-pile-top}
\begin{tabular}{lrrrr}
\toprule
Variant & SAT & UNSAT & Geomean speedup & Max speedup \\
\midrule
$K = 25$ (most aggressive)              & \num{26} & \num{23} & \num{1103}$\times$ & \num{4547}$\times$ \\
$K = 30$ $+$ king-pin                   & \num{18} & \num{31} & \num{225}$\times$  & \num{10470}$\times$ \\
$K = 35$ (least aggressive)             & \num{39} & \num{10} & \num{10}$\times$   & \num{51}$\times$ \\
\bottomrule
\end{tabular}
\end{table}

\subsection{Hardened Social Golfers}\label{sec:eval-hardened-sg}

Social Golfers~\cite{gent1999csplib} asks for a weekly schedule
of $g = n_{\text{pg}}
\cdot n_{\text{gr}}$ golfers ($n_{\text{pg}}$ per group,
$n_{\text{gr}}$ groups) into $n_{\text{rounds}}$ weeks, such that
no two golfers share a group more than once.  We harden the StreamLLM-distributed model with three
literature-standard additions: a week-1 canonical partition,
a golfer-1 group anchor, and an inter-week lex ordering
(full constraint listings in \Cref{app:hardening}).
Together these collapse the standard SG
symmetry group on the deployed baseline.  Even with this
hardening in place, \num{61} of the test instances drawn from
\cite{voboril2024streamllm} and \cite{spracklen2023portfolios}
remain solvable but non-trivially hard for the baseline (median
solve time on the order of minutes; total baseline wall-clock
across the \num{61}-instance set is \num{19.0}\,h).

\paragraph{Results.}
Our pipeline produces a pool of \num{23} candidate streamliners
on hardened SG (\num{11} from the LLM-statistics path, \num{12}
from the LLM-discovery path).  The deployed family-budget
portfolio achieves \num{98.6}\% portfolio time savings on the
\num{61}-instance test set, essentially matching the
\num{98.7}\% pool ceiling (within \num{0.1}\,pp).  The
simple-top-3 reading (comparison baseline) reaches \num{94.4}\%;
the \num{4.2}-percentage-point improvement of family-budget
over simple top-3 comes from including a second variant of the
\verb|sg_lex| family in the deployed portfolio.  The best
single streamliner achieves a geometric-mean speedup of
\num{356}$\times$ over the \num{18} of \num{61} instances on
which it retains satisfiability, with a per-instance maximum
of \num{2784}$\times$.

\paragraph{Statistics-path picks.}
The two statistics-path picks are both beyond-hardening
canonicalisations:
\begin{enumerate}
\item \emph{(LLM-statistics)} a closed-form algebraic
assignment for week 2:
\begin{lstlisting}
forall(g in Golfer)(assign[g, 2] =
  ((((g-1) div n_per_group)
    + ((g-1) mod n_per_group)) mod n_groups) + 1)
\end{lstlisting}
Filter \texttt{features.3[4]} correlates at $|r| = 0.97$
with \emph{subsquare\_2x2\_sum\_variance}, the variance of
$2\times2$ block sums in the scalar encoding, which is
near-constant across all \num{2000} solutions (std $= 0.14$).
This gives \emph{subsquare\_2x2\_sum\_variance} the highest
CNN endorsement among all \num{30} properties, directing the
LLM's attention to $2\times2$ block structure as the
load-bearing signal.  With this framing, the LLM inspects the
raw sample solutions, observes that week-2 assignments follow
a specific cyclic round-robin pattern relative to week 1, and
derives the algebraic formula.  The CNN's role here is to act
as an attention mechanism, elevating the right structural
feature above the remaining properties so the LLM focuses on
the correct part of the solution structure.
\item \emph{(LLM-statistics)} a per-golfer week-to-week change
constraint excluding golfer 1:
\begin{lstlisting}
forall(g in Golfer where g > 1, w in 1..n_rounds-1)
  (assign[g, w] != assign[g, w+1])
\end{lstlisting}
This forbids any non-canonicalised golfer from being assigned
to the same group on two consecutive weeks.  The stats path
surfaced it via filter \texttt{features.3[32]}, which
correlates at $|r| = 0.85$ with the
\emph{horizontal\_adjacency\_diff} property (a measure of
how much consecutive-week group assignments differ across the
solution).  That property is near-constant with mean $1.32$
and std $0.009$, indicating golfers almost never repeat a
group in consecutive weeks in valid solutions; the LLM reads
this near-constant signal and emits the constraint.
Validation confirms this is a genuine streamliner: it retains
satisfiability on \num{55} of \num{61} test instances
(with \num{6} rendered unsatisfiable), confirming that valid
solutions exist in which a golfer does occupy the same group
in consecutive weeks.
\end{enumerate}

Both constraints are beyond-hardening canonicalisations that
impose additional structure on top of the standard hardening
without being among the canonical lex-leader, value-precedence,
or week-permutation symmetry breaks; they are constraints the
standard hardening does not capture.

\paragraph{What the CNN sees.}
Unlike Black Hole and Vessel Loading, Social Golfers produces
nearly uniform activation maps with no diagonal striping or
geometric silhouette.  The structural patterns the CNN detects
(consistent $2\times2$ block sums and horizontal edge
regularity between consecutive weeks) hold uniformly across
the entire canvas, so filters fire at similar strength
everywhere rather than at identifiable positions.  The CNN
found real patterns, but they are diffuse rather than
localised, making spatial hotspot data less informative here
than for BH or VL.

For the discovery path, diffuse activations give the LLM less
positional evidence to anchor on, and it falls back on
prior-driven canonicalisation drawn from its knowledge of SG
group-permutation structure rather than reading a
spatially-concentrated CNN signal.  Where activations are
spatially concentrated (BH, VL), the discovery path extracts
CNN-grounded geometric constraints
(\Cref{sec:eval-hardened-vl}); where they are flat, as here,
the CNN's contribution is through the stats-path correlation
ranking rather than spatial signal.

\subsection{Hardened Vessel Loading}\label{sec:eval-hardened-vl}

Vessel Loading~\cite{gent1999csplib} is a 2D
rectangular-packing problem: place
\verb|n_containers| rectangles on a \verb|deck_width| by
\verb|deck_length| deck such that containers of incompatible
classes respect minimum separation distances.  We harden the StreamLLM-distributed model with two additions:
a lex-leader symmetry break over identical containers, and a
square-container orientation canonicalisation (full constraint
listings in \Cref{app:hardening}).  Because the baseline VL
model permits container rotation, this
hardening is more substantial than the symmetry break on Social
Golfers and substantially compresses the baseline solve-time
distribution.  Even after hardening, however, the
\num{22}-instance composite test set retains a meaningful
distribution of harder cases drawn from
\cite{spracklen2023portfolios} (total baseline wall-clock
\num{6.1}\,h).

\begin{figure}[!htbp]
\centering
\includegraphics[width=\textwidth]{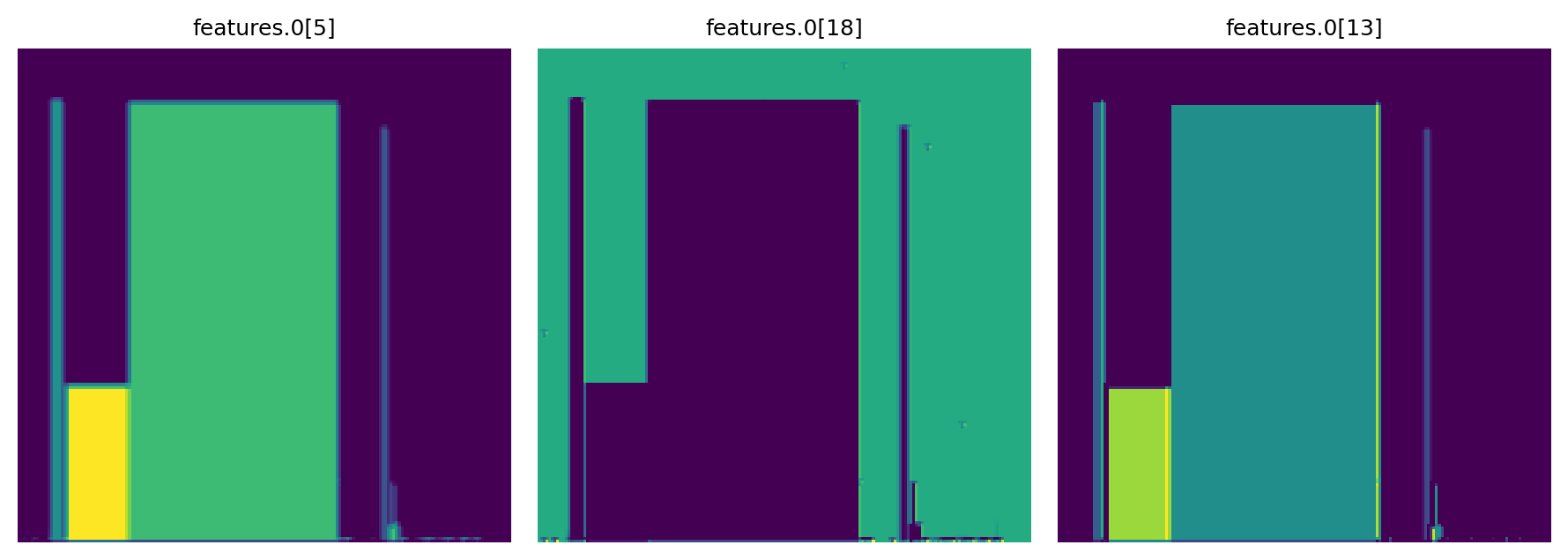}
\caption{Mean activation maps for the three highest-variance
\texttt{features.0} filters trained contrastively on Vessel
Loading solutions versus packing near-miss negatives (training
instance \texttt{0051e4a17e\ldots}; mean over the
enumerated-solutions corpus).  Columns = \texttt{Left}
coordinate ($x$-axis of deck); rows = \texttt{Bottom}
coordinate ($y$-axis); colour encodes activation strength
(dark purple = none, teal/green = moderate activation over
the main deck region, yellow = peak).  Across most filters the
same silhouette recurs: vertical bands at the deck walls,
a dominant interior region, and a consistent bright
bottom-left hot-spot.  The bottom-left hot-spot is the spatial
signal the LLM-discovery path converts into the
packing-heuristic streamliners (\texttt{Bottom[c] mod 2 = 0},
same-class \texttt{Left} clustering) representative of the
discovery-path pool.}
\label{fig:cnn-activations-vl}
\end{figure}

\paragraph{What the CNN sees.}
\Cref{fig:cnn-activations-vl} renders the learned filter
activations on a representative VL training instance.  VL
solutions are rasterised to a \texttt{deck\_length}
$\times$ \texttt{deck\_width} grid where each cell holds the
class ID of the container occupying that position (zero for
empty); the CNN therefore sees a spatial map of the deck.
In each activation panel, columns correspond to the
$x$-axis (\texttt{Left} coordinate, left deck wall at
column~$0$), rows to the $y$-axis (\texttt{Bottom}
coordinate), and colour encodes filter response
strength (dark purple = no activation,
teal/green = moderate activation over the main deck region,
bright yellow = peak activation).  Across the
top-variance \texttt{features.0} filters the same silhouette
recurs: thin vertical bands at the deck walls, a dominant
interior region, and a consistent bright bottom-left hot-spot.
The discovery-path picks are precisely the constraints this
geometry would suggest: a global rotation suppression keying
on the canonical orientation visible in the silhouette,
and a same-class horizontal-clustering bound keying on the
bottom-left hot-spot's locality.  The LLM is reading the
activation maps in this case rather than relying on
problem-description priors alone: what the CNN exposes
spatially, the LLM converts into a packing heuristic.

\paragraph{Results.}
Our pipeline produces a pool of \num{40} candidate streamliners
on hardened VL (\num{23} from the LLM-statistics path, \num{17}
from the LLM-discovery path).  The deployed family-budget
portfolio achieves \num{98.8}\% portfolio time savings on the
\num{22}-instance test set, recovering most of the way to the
\num{99.6}\% pool ceiling.  The simple-top-3 reading
reaches \num{97.4}\%, already close to the pool ceiling on this problem;
the family-budget improvement is modest (\num{+1.4}\,pp) because
VL's pool is already near-saturated under simple-top-3.  The
best single streamliner achieves a geometric-mean speedup of
\num{932}$\times$ over the \num{17} of \num{22} instances on
which it retains satisfiability, with a maximum per-instance
speedup of \num{8389}$\times$.

\paragraph{Illustrative pool picks.}
Three constraints from different families illustrate the
range of structure the pipeline discovers, mixing both LLM paths:
\begin{enumerate}
\item \emph{(LLM-statistics)} an orientation-aware left-edge
bound on containers smaller than the deck:
\begin{lstlisting}
forall(c in Containers
       where max(width[c], length[c]) <= deck_width)
  (Left[c] <= deck_width - max(width[c], length[c]))
\end{lstlisting}
This enforces that any container that fits within the deck
width in either orientation must be placed so that its left
edge plus its larger dimension does not exceed the deck width,
a packing-tightness constraint that handles both rotations.
The CNN signal came from the \emph{mean\_Left\_all} and
\emph{std\_Left\_all} properties ($|r|>0.94$ across all
three ensemble seeds), indicating that Left coordinates are
tightly constrained in valid solutions.  The LLM uses this
signal as a cue and derives the orientation-aware formula from
the model structure: neither alone produces the constraint.
It retains satisfiability on \num{18} of \num{22} test
instances with a geometric-mean speedup of \num{439}$\times$.
\item \emph{(LLM-statistics)} a bottom-edge grid-alignment
constraint:
\begin{lstlisting}
forall(c in Containers)(Bottom[c] mod 2 = 0)
\end{lstlisting}
This requires every container's bottom edge to land on an even
$y$-coordinate.  The constraint is not in the model and cannot
be derived from reading it; it is an emergent property of the
specific container dimensions in these instances.  When
containers with even-valued heights are packed starting from
$y=0$, each successive bottom edge naturally lands on an even
coordinate, and the CNN detects this regularity via the
\emph{n\_boundaries} property (|r| = 0.97), which is
near-constant across all solutions.  The LLM translates
``container edges consistently align to even grid positions''
into the mod-2 constraint.  By halving the candidate
$y$-coordinates for every container simultaneously, it prunes
the search space dramatically: \num{16} of \num{22} test
instances retain satisfiability with a geometric-mean speedup
of \num{646}$\times$.
\item \emph{(LLM-discovery)} a same-class horizontal clustering
constraint:
\begin{lstlisting}
forall(c, k in Containers
       where c < k /\ class[c] = class[k])
  (abs(Left[c] - Left[k]) <= deck_width div 3)
\end{lstlisting}
The CNN's activation maps expose a spatial regularity: in
high-activation solutions, containers sharing the same cargo
class are consistently co-located in the same region of the
deck rather than spread across it.  The LLM reads this
clustering signal and translates it into a proximity bound:
any two same-class containers must have their left edges
within one-third of the deck width of each other.  This
complements the symmetry break on \emph{identical} containers
(which requires matching width and length as well as class)
by restricting the horizontal spread of same-class containers
regardless of their dimensions.  It retains satisfiability on
\num{18} of \num{22} test instances with a geometric-mean
speedup of \num{654}$\times$.  Notably, the LLM generates
this bound without referencing the \texttt{separation} matrix
already in the model, which specifies the minimum required
distance between containers of each class pair.  A more
precise version could derive a bound consistent with those
separation values; this illustrates the prompt-improvement
direction discussed in \Cref{sec:conclusion}.
\end{enumerate}

The hardened VL winners are \emph{structural packing
heuristics} rather than further canonicalisations: they restrict
where containers are placed and how their classes
relate, rather than canonicalising the symmetry orbit of the
hardened baseline.  This is a different kind of structure from
the Social Golfers winners and reflects VL's geometric, rather
than symmetric, residual structure.

\section{Analysis}\label{sec:analysis}

\subsection{Which LLM path carries the headline?}\label{sec:analysis-source}

\begin{figure}[!htbp]
\centering
\includegraphics[width=\textwidth]{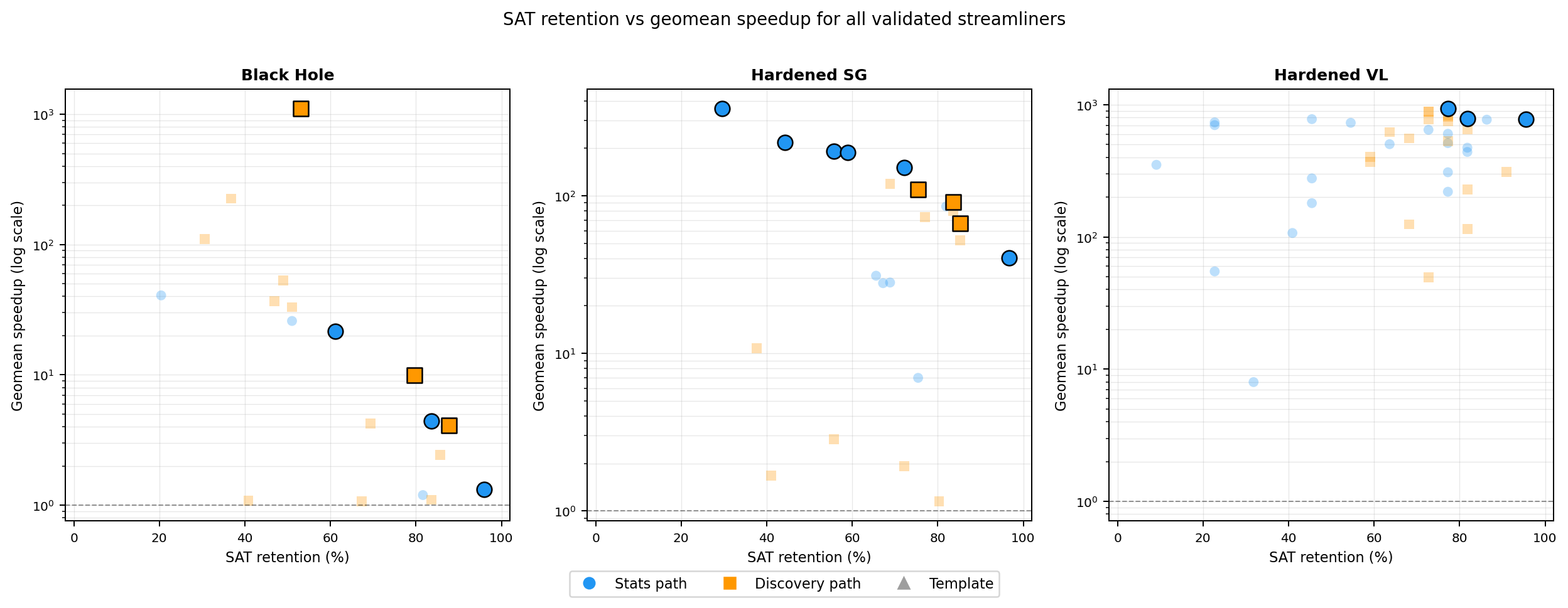}
\caption{SAT retention (\% of test instances solved) vs
geomean speedup on retained instances (log scale) for all
validated streamliners across the three problems.  Blue
circles = \texttt{llm\_stats} path; orange squares =
\texttt{llm\_discovery} path.  No template candidates appear
above $1\times$ speedup on the hardened evaluation regime.
The dashed line marks $1\times$ (no improvement).}
\label{fig:pareto-front}
\end{figure}

A natural question given the two-path design is whether one
path dominates the other.  \Cref{fig:pareto-front} addresses
this by plotting every validated streamliner as a point on
the SAT-retention vs geomean-speedup plane.

Two observations stand out.  First, SAT retention and speedup
are in fundamental tension: tighter constraints prune the
search space more aggressively and deliver higher per-instance
speedup where they apply, but render more instances
unsatisfiable.  This tradeoff is visible in all three panels
as a downward-sloping Pareto frontier from top-left (high
speedup, low retention) to bottom-right (lower speedup, broad
coverage).  A useful streamliner portfolio needs both: a
high-speedup/low-retention member to win on the instances it
retains, and a conservative member to cover the rest.

Second, the two LLM paths occupy different regions of the
tradeoff space.  On Black Hole the discovery path dominates
the high-speedup end (the pile-top family), while on hardened
SG the stats path owns the top-left.  On VL both paths are
competitive throughout.  The two paths address different
aspects of the solution-set signal: the \texttt{llm\_stats}
path is strongest when structural regularity is captured by
the per-property library, while \texttt{llm\_discovery} is
strongest when the regularity involves variables outside that
library (the BH pile-top coordinates, the VL grid-alignment
pattern).  Templates appear in the BH pool but contribute
nothing above $1\times$ on the hardened regime and are
omitted from the figure.

\subsection{Selection inefficiency and the role of the pool}\label{sec:analysis-selection}

As \Cref{tab:results-combined} shows, hardened SG/VL
exhibit small selection gaps ($\leq 5$\,pp) while Black Hole
exhibits a \num{15.2}-percentage-point gap.  We can decompose
the gap further by examining the candidate constraint
\emph{families} present in each pool.

On hardened SG and VL, the deployed picks span complementary
constraint families, each targeting a distinct structural
regularity.  The 4-way race in the portfolio rarely sees two
streamliners competing on the same instance, and the static
rule recovers most of the pool ceiling.

On Black Hole, by contrast, the simple top-3 picks are all from
the same \emph{family}: pile-top $y$-coordinate bounds at three
aggressiveness levels.  When this family is the per-instance
optimum the portfolio captures it, but for instances where the
optimum lies in a different family the portfolio falls back to
baseline.  This is the structural origin of BH's
\num{15.2}\,pp selection gap, which the family-budget
allocation (\Cref{sec:pipeline-portfolio}) is designed to
address.  As \Cref{tab:results-combined} reports, family-budget
recovers \num{9.7} of the \num{15.2}\,pp gap, leaving a
\num{5.5}\,pp residual.  This instance-level variability is
directly visible in \Cref{fig:speedup-heatmap}: the BH panel
shows a highly fragmented pattern where no single streamliner
dominates across the full instance set, in contrast to the more
consistent coverage visible in the SG and VL panels.

\subsection{Sensitivity of family-budget to $k$ and $m$, and wall-clock vs.\ CPU cost}\label{sec:analysis-family-sensitivity}

The family-budget mechanism (\Cref{sec:pipeline-portfolio}) has
two parameters: the number of families $k$ deployed and the
number of members $m$ retained per family.  Throughout the
evaluation we fix $k = 3$ for direct comparability with prior
portfolio work, but here we sweep both dimensions and
additionally report CPU-adjusted savings alongside wall-clock
savings to characterise the compute cost of the parallel race.

\paragraph{Intra-family count $m$.}
\Cref{tab:family-budget-sweep} reports wall-clock portfolio
savings at $k = 3$ for $m \in \{1, 2, 3\}$ and for
``all members'' (no intra-family cap).  Three observations.
First, the diminishing-returns prediction is borne out
empirically: $m = 3$ captures essentially all of the
recoverable gain on each problem, and the marginal gain from
$m = 2$ to $m = 3$ is already smaller than from $m = 1$ to
$m = 2$ on Black Hole.  Second, growing $m$ beyond three
members per family \emph{degrades} performance on Vessel
Loading: VL's \verb|vl_no_rotation| family has eight variants
whose later members are weak, and including them all causes
slot dilution that prevents the strong members from completing
within their shrunken slot.  This corroborates the mechanism's
design hypothesis: the value of intra-family diversity
saturates early, and exceeding the saturation point hurts.
Third, $m = 1$ reduces the family-budget portfolio to one
member-per-family (effectively a family-balanced top-$k$
ranking with full per-streamliner time) and recovers only a
small fraction of the family-budget gain on SG and VL
(\num{+0.9}\,pp and \num{+0.2}\,pp over simple top-3
respectively), and is slightly \emph{worse} than simple top-3
on Black Hole (\num{-0.5}\,pp) because the single best pile-top
member per family cannot replicate the slot-sharing benefit.
The family-aware \emph{allocation}
(slot-shared compute across multiple members) does the bulk of
the work; family-aware \emph{selection} alone is insufficient.

\begin{table}[t]
\centering
\caption{Sensitivity to intra-family member count $m$ at
$k = 3$ families (wall-clock savings).  Best $m$ per problem in bold;
``all members'' = no cap on intra-family size.  All values
simulated from the shared validation-pool data used for the
\Cref{sec:eval} headlines.}
\label{tab:family-budget-sweep}
\begin{tabular}{lrrrr}
\toprule
              & $m = 1$ & $m = 2$ & $m = 3$ & all members \\
\midrule
Hardened SG   & \num{95.3}\%  & \num{95.2}\%  & \textbf{\num{98.6}\%} & \num{98.7}\% \\
Hardened VL   & \num{97.6}\%  & \num{98.8}\%  & \textbf{\num{98.8}\%} & \num{95.3}\% \\
Black Hole    & \num{79.2}\%  & \num{89.4}\%  & \textbf{\num{89.4}\%} & \num{83.9}\% \\
\bottomrule
\end{tabular}
\end{table}

\paragraph{Wall-clock vs.\ CPU-adjusted savings.}
For Hardened SG and Hardened VL the CPU-adjusted figures track
wall-clock savings closely across all $(k, m)$ settings (within
\num{4}\,pp), because the winning streamliners solve their
instances so quickly that $(k+1) \times t_{\rm winner}$ remains
a small fraction of $t_b$.  At the paper's default $(k=3,
m=3)$: SG \num{94.5}\% CPU-adjusted (vs.\ \num{98.6}\%
wall-clock); VL \num{95.1}\% (vs.\ \num{98.8}\%).

Black Hole shows a larger gap (\num{89.4}\% wall-clock,
\num{57.7}\% CPU-adjusted at $k=3$, $m=3$) because BH
streamliner SAT coverage is lower: a non-trivial fraction of
instances are solved only by the baseline, so the parallel
processes that fail on those instances consume their full slot
budget without contributing.  The wasted compute accumulates
in the CPU accounting but is invisible in the wall-clock figure,
which records only the winner's time.

\Cref{tab:bh-km-sweep} shows the full $(k, m)$ sweep for Black
Hole, reporting both wall-clock savings (WC) and CPU-adjusted
savings (CPU).  Increasing $k$ improves wall-clock monotonically
(more diverse families cover more instances) but \emph{degrades}
CPU-adjusted savings at fixed $m = 1$, because each additional
family adds another parallel process that frequently exhausts
its slot without finding SAT.  The most CPU-efficient settings
are low-$k$, moderate-$m$: $(k=2, m=2)$ delivers \num{88.8}\%
wall-clock at \num{66.4}\% CPU-adjusted, the best CPU figure in
the table.  The paper's default $(k=3, m=3)$ trades
\num{8.7}\,pp of CPU efficiency for \num{0.6}\,pp of additional
wall-clock savings over $(k=2, m=2)$, a reasonable exchange
when spare cores are available.  Practitioners under strict
compute budgets may prefer the $(k=2, m=2)$ configuration.

\begin{table}[t]
\centering
\caption{Black Hole portfolio savings across family count $k$
and per-family member count $m$.  Each cell shows wall-clock
savings (WC) / CPU-adjusted savings (CPU).  Paper default
$(k=3, m=3)$ marked $\dagger$; oracle pool ceiling is
\num{94.9}\% wall-clock.}
\label{tab:bh-km-sweep}
\begin{tabular}{lrrr}
\toprule
 & $m = 1$ & $m = 2$ & $m = 3$ \\
\midrule
$k = 1$ & \num{54.7}\% / \num{9.4}\%   & \num{77.8}\% / \num{55.6}\%  & \num{78.0}\% / \num{56.0}\% \\
$k = 2$ & \num{75.2}\% / \num{25.6}\%  & \num{88.8}\% / \num{66.4}\%  & \num{88.8}\% / \num{66.5}\% \\
$k = 3$ & \num{79.2}\% / \num{16.9}\%  & \num{89.4}\% / \num{57.7}\%  & \num{89.4}\% / \num{57.7}\%$^\dagger$ \\
$k = 4$ & \num{81.6}\% / \num{8.2}\%   & \num{90.0}\% / \num{49.8}\%  & \num{90.0}\% / \num{49.9}\% \\
$k = 5$ & \num{85.0}\% / \num{10.2}\%  & \num{90.0}\% / \num{40.1}\%  & \num{90.0}\% / \num{40.2}\% \\
\bottomrule
\end{tabular}
\end{table}

\section{Limitations and Threats to Validity}\label{sec:threats}

\paragraph{Breadth of direct comparison.}
We report a direct head-to-head against
StreamLLM~\cite{voboril2024streamllm} only on Black Hole, the
one problem where standard hardening is not applicable and the
two pipelines therefore face the same task.  For Social Golfers
and Vessel Loading we evaluate on \emph{hardened} variants of
the StreamLLM-distributed models, which is a different
evaluation regime; a reader cannot directly compare our hardened
SG and VL numbers against StreamLLM's reported figures because
the underlying baselines differ.  We have not re-run the
StreamLLM pipeline against our hardened models.  A
hardened-baseline rerun of StreamLLM, while methodologically
clean, is computationally expensive and beyond the scope of
this paper; we leave it to future work.

\paragraph{LLM dependence.}
Constraint synthesis uses Claude Opus as the prompted LLM
throughout.  We did experiment with self-hosted open-weight
models via Ollama (specifically the
\texttt{qwen2.5-coder:14b} class of models, the largest size
our available hardware could run), but the
generated MiniZinc was not syntactically reliable enough to
sustain the constraint-synthesis pass, and we were unable to
host larger models that might have done better.  An
open-weight reproduction with sufficiently capable hardware is
therefore left to future work; it is desirable both for cost
reasons (the headline pools are independent of solver runs)
and for reproducibility under proprietary-model discontinuation.
We report the LLM client and version in each experiment's
\texttt{experiment\_config.json}.

\paragraph{Run-to-run variance and reproducibility.}
Single-seed pipeline runs have non-trivial variance because
both the CNN training and the LLM constraint synthesis are
non-deterministic.  We mitigate the CNN side with an ensemble
of $K=3$ random seeds per training instance; the LLM side is
mitigated by the semantic-clustering pass that absorbs
parametrically-equivalent variants into representative
clusters.  Headline numbers in this paper come from a single
end-to-end pipeline run per problem; we have not run multiple
end-to-end seeds and report variance bands.  This is a real
limitation: a reviewer cannot infer from our numbers how
sensitive the deployed top-3 picks are to the seed used for
constraint synthesis.

\paragraph{CNN ablation and architecture.}
The meaningful published alternative to CNN-grounded synthesis
is LLM-only generation from the model text (StreamLLM), which
we compare against directly on Black Hole (\Cref{sec:eval-bh}).
The mechanical template path is the other natural reference and
is already in the pipeline; it contributes no headline picks
(\Cref{fig:pareto-front}).  The BH pile-top constraint
is unreachable by any fixed-property-library approach
(\Cref{sec:eval-bh}), establishing the CNN's contribution by
construction on that problem.  The one open question is
whether the CNN correlation ranking on the \texttt{llm\_stats}
path adds value over simply presenting all properties
unranked; we leave this to future work.  A related question
is network depth: in practice the most interpretable signal
was consistently in the first layer, and whether the
\num{3}-layer architecture is necessary over a single
convolutional layer has not been tested.

\paragraph{Variance check.}
The pipeline implicitly assumes that the property library
contains at least one property with non-zero per-solution
variance and bounded range; without such a property no
streamliner candidate can be generated by the
\texttt{llm\_stats} path.  Problems whose feasible solutions
share all aggregate properties (e.g., Steiner Triple Systems,
where every feasible solution has identical row sums, column
sums, and pair counts) will yield no useful candidates from
the statistics path, though the discovery path may still
identify structure from the filter-contrast pairs.  We have
not evaluated such problems in this paper.

\section{Conclusion}\label{sec:conclusion}

We have presented a pipeline for streamliner synthesis driven by
CNN pattern recognition on enumerated solutions.  The central
idea is to use the CNN as a grounding mechanism: rather than
asking the LLM to reason from model text alone (as in
StreamLLM) or from raw solution arrays (impractical at scale),
the CNN first identifies which structural properties of the
solution distribution are most discriminative, and then directs
the LLM's attention to that observed evidence.  The result is
constraint synthesis that is both unconstrained in shape (the
LLM can propose any MiniZinc expression) and anchored in
structure that actually holds in the feasible solution set
rather than inferred from problem description alone.
Per training instance, a Convolutional Neural Network is
trained contrastively to detect structural patterns the
feasibility predicate exploits, and its discriminative signal
is surfaced to a Large Language Model along two paths:
filter--property correlation rankings alongside per-property
statistics, and direct filter-contrast on high-activation and
low-activation solution pairs.  The LLM translates the
resulting structural evidence into candidate MiniZinc
streamliners, which are pooled, clustered, and validated
against cached baselines.

We evaluated the pipeline in the regime where automated
streamliner discovery is genuinely needed: hardened benchmark
models with literature-standard symmetry-breaking and implied
constraints already in place, plus Black Hole, the one
benchmark in our suite without a canonical hardening.
Deployed at $k = 3$ families with $m = 3$ members per family,
the family-budget portfolio delivers \num{98.8}\% time
reduction on hardened Vessel Loading with a
best-single-streamliner geometric-mean speedup of
\num{932}$\times$; \num{98.6}\% on hardened Social Golfers
with a \num{356}$\times$ best-single geomean; and
\num{89.4}\% on Black Hole with a \num{1103}$\times$
best-single geomean.  On Black Hole, where direct head-to-head
comparison with prior work is methodologically clean, the
\num{89.4}\% family-budget result exceeds StreamLLM's
realtime configuration ($\approx$\num{80}\%); the
simple-top-3 reading (\num{79.7}\%) matches that prior work
for like-for-like comparison, and family-budget closes
\num{9.7} of the \num{15.2}-percentage-point pool ceiling
gap with no additional compute envelope.

Across all three problems, both LLM paths contribute
headline picks; the CNN's filter-contrast path is
load-bearing on Black Hole specifically, where it surfaces
constraints over layout coordinates that the per-property
statistics path does not reach.  The discovered streamliners
are structurally distinct from what the standard hardening
provides: class-based packing constraints on Vessel
Loading, beyond-hardening canonicalisations on Social
Golfers, and layout-coordinate bounds on Black Hole.

A secondary observation is that the CNN's activation maps
carry diagnostic information about which kind of constraint
the discovery path produces.  On Vessel Loading and Black
Hole the activations expose interpretable spatial structure
(deck silhouette and per-pile invariants respectively) and the
discovery picks key directly on that structure; the Black Hole
pile-top constraint is structurally unreachable by any
fixed-property-library approach, providing direct evidence of
the CNN's contribution on that problem.  On Social Golfers
the activations are flat and the discovery path falls back on
prior-driven canonicalisations.

The most natural direction for future work is richer CNN
training: the current binary contrastive objective converges
on the single most discriminative pattern per instance, and
harder negatives or multi-property auxiliary tasks would
encourage filter specialisation across independent structural
signals.  A second direction is prompt refinement: although
the LLM already receives the full MiniZinc model alongside
the CNN evidence, it does not always reason carefully about
consistency with existing model constraints when proposing
bounds.  Explicitly instructing the LLM to verify that a
proposed constraint does not conflict with or duplicate
constraints already in the model would improve the
precision of generated candidates and reduce candidates that
are trivially implied or structurally inconsistent.  A third
direction is finer family tagging for the portfolio
(constraint-AST clustering or behavioural similarity rather
than descriptor prefix) to recover the residual gap to the
full-pool oracle.  A broader open-weight LLM reproduction
would also strengthen reproducibility.

\appendix

\section{Hardening constraints}\label{app:hardening}

\subsection*{Social Golfers}

We add three literature-standard constraints~\cite{crawford1996symbreaking} to
the StreamLLM-distributed Social Golfers model:

\begin{itemize}
\item \emph{Week-1 canonical partition.} Fix the entire first-week column to a
canonical assignment, breaking golfer-permutation symmetry within week 1:
\begin{lstlisting}
forall(g in Golfer)(assign[g, 1] = ((g-1) div n_per_group) + 1)
\end{lstlisting}
\item \emph{Golfer-1 group anchor.} Fix golfer~1 to group~1 in every week
beyond the first, naming ``group~1'' as the group containing golfer~1:
\begin{lstlisting}
forall(w in Week where w > 1)(assign[1, w] = 1)
\end{lstlisting}
\item \emph{Inter-week lex ordering.} Impose lex order on weeks
2\ldots$n_{\rm rounds}$~\cite{flener2002rowcolsym}, breaking week-permutation
symmetry (week~1 is excluded as its column is already canonicalised).
\end{itemize}

\subsection*{Vessel Loading}

We add two constraints~\cite{crawford1996symbreaking} to the
StreamLLM-distributed Vessel Loading model:

\begin{itemize}
\item \emph{Lex-leader symmetry break over identical containers.} Containers
sharing width, length, and class are interchangeable; their
(\texttt{Left}, \texttt{Bottom}) pairs are forced into lex order:
\begin{lstlisting}
forall(c1, c2 in Containers
       where c1 < c2 /\ width[c1]=width[c2]
                    /\ length[c1]=length[c2]
                    /\ class[c1]=class[c2])
  (lex_lesseq([Left[c1], Bottom[c1]],
              [Left[c2], Bottom[c2]]))
\end{lstlisting}
\item \emph{Square-container orientation canonicalisation.} Containers with
$\textit{width} = \textit{length}$ have their rotation flag fixed, eliminating
the trivial rotation symmetry that would otherwise double-count square
placements:
\begin{lstlisting}
forall(c in Containers where width[c]=length[c])
  (not rotated[c])
\end{lstlisting}
\end{itemize}

\bibliographystyle{plain}
\bibliography{references}

@article{roussel2011runsolver,
  author    = {Roussel, Olivier},
  title     = {Controlling a Solver Execution: the \texttt{runsolver} Tool},
  journal   = {Journal on Satisfiability, Boolean Modeling and Computation},
  volume    = {7},
  number    = {4},
  pages     = {139--144},
  year      = {2011},
}

@article{bessiere2017acquisition,
  author    = {Bessiere, Christian and Coletta, R{\'{e}}mi and Hebrard, Emmanuel and Katsirelos, George and Lazaar, Nadjib and Narodytska, Nina and Quimper, Claude-Guy and Walsh, Toby},
  title     = {Constraint acquisition},
  journal   = {Artificial Intelligence},
  volume    = {244},
  pages     = {315--342},
  year      = {2017},
  doi       = {10.1016/j.artint.2015.08.001},
}

@book{rossi2006handbook,
  editor    = {Rossi, Francesca and van Beek, Peter and Walsh, Toby},
  title     = {Handbook of Constraint Programming},
  publisher = {Elsevier},
  year      = {2006},
  series    = {Foundations of Artificial Intelligence},
}

@inproceedings{crawford1996symbreaking,
  author    = {Crawford, James and Ginsberg, Matthew and Luks, Eugene and Roy, Amitabha},
  title     = {Symmetry-breaking predicates for search problems},
  booktitle = {Knowledge Representation and Reasoning (KR)},
  year      = {1996},
}

@inproceedings{gomes2004streamlined,
  author    = {Gomes, Carla P. and Sellmann, Meinolf},
  title     = {Streamlined Constraint Reasoning},
  booktitle = {Principles and Practice of Constraint Programming (CP)},
  year      = {2004},
}

@inproceedings{spracklen2018montecarlo,
  author    = {Spracklen, Patrick and Akg{\"u}n, {\"O}zg{\"u}r and Miguel, Ian},
  title     = {Monte Carlo Tree Search on a Model Lattice: Automatic Streamliner Generation},
  booktitle = {Principles and Practice of Constraint Programming (CP)},
  year      = {2018},
}

@article{spracklen2023portfolios,
  author    = {Spracklen, Patrick and Dang, Nguyen and Akg{\"u}n, {\"O}zg{\"u}r and Miguel, Ian},
  title     = {Automated streamliner portfolios for constraint satisfaction problems},
  journal   = {Artificial Intelligence},
  volume    = {319},
  pages     = {103915},
  year      = {2023},
  publisher = {Elsevier},
  doi       = {10.1016/j.artint.2023.103915},
  url       = {https://www.sciencedirect.com/science/article/pii/S0004370223000619},
  note      = {Benchmark instances and code: \url{https://github.com/stacs-cp/automated-streamliner-portfolios}},
}

@article{voboril2024streamllm,
  author    = {Voboril, Florentina and {Peruvemba Ramaswamy}, Vaidyanathan and Szeider, Stefan},
  title     = {Generating Streamlining Constraints with Large Language Models},
  journal   = {Journal of Artificial Intelligence Research},
  volume    = {84},
  pages     = {16:1--16:19},
  year      = {2025},
  doi       = {10.1613/jair.1.18965},
  note      = {Code and instances: \url{https://zenodo.org/records/14757597}},
}

@inproceedings{beldiceanu2012modelseeker,
  author    = {Beldiceanu, Nicolas and Simonis, Helmut},
  title     = {A Model Seeker: Extracting Global Constraint Models from Positive Examples},
  booktitle = {Principles and Practice of Constraint Programming (CP)},
  year      = {2012},
}

@inproceedings{bau2017netdissect,
  author    = {Bau, David and Zhou, Bolei and Khosla, Aditya and Oliva, Aude and Torralba, Antonio},
  title     = {Network Dissection: Quantifying Interpretability of Deep Visual Representations},
  booktitle = {Conference on Computer Vision and Pattern Recognition (CVPR)},
  year      = {2017},
}

@inproceedings{wang2019satnet,
  author    = {Wang, Po-Wei and Donti, Priya and Wilder, Bryan and Kolter, J.~Zico},
  title     = {{SATNet}: Bridging Deep Learning and Logical Reasoning Using a Differentiable Satisfiability Solver},
  booktitle = {International Conference on Machine Learning (ICML)},
  year      = {2019},
}

@article{selsam2019neurosat,
  author  = {Selsam, Daniel and Lamm, Matthew and B{\"u}nz, Benedikt and Liang, Percy and de Moura, Leonardo and Dill, David L.},
  title   = {Learning a {SAT} Solver from Single-Bit Supervision},
  journal = {International Conference on Learning Representations (ICLR)},
  year    = {2019},
}

@article{frisch2008essence,
  author    = {Frisch, Alan M. and Harvey, Warwick and Jefferson, Christopher and Mart{\'\i}nez Hern{\'a}ndez, Bernadette and Miguel, Ian},
  title     = {{Essence}: A constraint language for specifying combinatorial problems},
  journal   = {Constraints},
  volume    = {13},
  number    = {3},
  pages     = {268--306},
  year      = {2008},
  publisher = {Springer},
  doi       = {10.1007/s10601-008-9047-y},
}

@inproceedings{nethercote2007minizinc,
  author    = {Nethercote, Nicholas and Stuckey, Peter J. and Becket, Ralph and Brand, Sebastian and Duck, Gregory J. and Tack, Guido},
  title     = {{MiniZinc}: Towards a Standard {CP} Modelling Language},
  booktitle = {Principles and Practice of Constraint Programming (CP)},
  year      = {2007},
}

@inproceedings{chu2018chuffed,
  author    = {Chu, Geoffrey},
  title     = {Chuffed: A Lazy Clause Generation Solver},
  year      = {2018},
  note      = {\url{https://github.com/chuffed/chuffed}},
}

@inproceedings{wetter2015streamlined,
  author    = {Wetter, James and Akg{\"u}n, {\"O}zg{\"u}r and Miguel, Ian},
  title     = {Automatically Generating Streamlined Constraint Models with {Essence} and {Conjure}},
  booktitle = {Principles and Practice of Constraint Programming (CP)},
  year      = {2015},
}

@article{akgun2022conjure,
  author    = {Akg{\"u}n, {\"O}zg{\"u}r and Frisch, Alan M. and Gent, Ian P. and Jefferson, Christopher and Miguel, Ian and Nightingale, Peter},
  title     = {{Conjure}: Automatic Generation of Constraint Models from Problem Specifications},
  journal   = {Artificial Intelligence},
  volume    = {310},
  pages     = {103751},
  year      = {2022},
  doi       = {10.1016/j.artint.2022.103751},
}

@inproceedings{spracklen2019optimisation,
  author    = {Spracklen, Patrick and Dang, Nguyen and Akg{\"u}n, {\"O}zg{\"u}r and Miguel, Ian},
  title     = {Automatic Streamlining for Constrained Optimisation},
  booktitle = {Principles and Practice of Constraint Programming (CP)},
  year      = {2019},
}

@inproceedings{charnley2006implied,
  author    = {Charnley, John and Colton, Simon and Miguel, Ian},
  title     = {Automatic Generation of Implied Constraints},
  booktitle = {European Conference on Artificial Intelligence (ECAI)},
  year      = {2006},
}

@inproceedings{lebras2012streamlined,
  author    = {Le Bras, Ronan and Gomes, Carla P. and Selman, Bart},
  title     = {From Streamlined Combinatorial Search to Efficient Constructive Procedures},
  booktitle = {AAAI Conference on Artificial Intelligence},
  year      = {2012},
}

@article{romera2023funsearch,
  author    = {Romera-Paredes, Bernardino and Barekatain, Mohammadamin and Novikov, Alexander and Balog, Matej and Kumar, M. Pawan and Dupont, Emilien and Ruiz, Francisco J. R. and Ellenberg, Jordan S. and Wang, Pengming and Fawzi, Omar and Kohli, Pushmeet and Fawzi, Alhussein},
  title     = {Mathematical Discoveries from Program Search with Large Language Models},
  journal   = {Nature},
  volume    = {625},
  pages     = {468--475},
  year      = {2024},
  doi       = {10.1038/s41586-023-06924-6},
}

@inproceedings{pei2023invariants,
  author    = {Pei, Kexin and Bieber, David and Shi, Kensen and Sutton, Charles and Yin, Pengcheng},
  title     = {Can Large Language Models Reason about Program Invariants?},
  booktitle = {International Conference on Machine Learning (ICML)},
  year      = {2023},
}

@inproceedings{wu2023lemur,
  author    = {Wu, Haoze and Barrett, Clark and Narodytska, Nina},
  title     = {{Lemur}: Integrating Large Language Models in Automated Program Verification},
  booktitle = {International Conference on Learning Representations (ICLR)},
  year      = {2024},
}

@inproceedings{smith2005streamlining,
  author    = {Smith, Casey and Gomes, Carla P. and Fern{\'a}ndez, C{\`e}sar},
  title     = {Streamlining Local Search for Spatially Balanced {Latin} Squares},
  booktitle = {International Joint Conference on Artificial Intelligence (IJCAI)},
  year      = {2005},
}

@inproceedings{flener2002rowcolsym,
  author    = {Flener, Pierre and Frisch, Alan M. and Hnich, Brahim and Kiziltan, Zeynep and Miguel, Ian and Pearson, Justin and Walsh, Toby},
  title     = {Breaking Row and Column Symmetries in Matrix Models},
  booktitle = {Principles and Practice of Constraint Programming (CP)},
  pages     = {462--477},
  year      = {2002},
}

@inproceedings{dang2022instancegen,
  author    = {Dang, Nguyen and Akg{\"u}n, {\"O}zg{\"u}r and Espasa, Joan and Miguel, Ian and Nightingale, Peter},
  title     = {A Framework for Generating Informative Benchmark Instances},
  booktitle = {Principles and Practice of Constraint Programming (CP)},
  volume    = {235},
  pages     = {18:1--18:18},
  year      = {2022},
  doi       = {10.4230/LIPIcs.CP.2022.18},
}

@misc{gent1999csplib,
  author       = {Gent, Ian P. and Walsh, Toby},
  title        = {{CSPLib}: A Benchmark Library for Constraints},
  howpublished = {\url{https://www.csplib.org/}},
  year         = {1999},
  note         = {Online benchmark library; problem 010 (Social Golfers), 008 (Vessel Loading), 023 (Black Hole patience)},
}

@inproceedings{gent2007blackhole,
  author    = {Gent, Ian P. and Jefferson, Christopher and Kelsey, Tom and Lynce, In{\^e}s and Miguel, Ian and Nightingale, Peter and Smith, Barbara M. and Tarim, S.~Armagan},
  title     = {Search in the Patience Game `{Black} {Hole}'},
  booktitle = {AI Communications},
  volume    = {20},
  number    = {3},
  pages     = {211--226},
  year      = {2007},
}

\end{document}